\documentclass{article}

     \PassOptionsToPackage{numbers, compress}{natbib}
\usepackage[final]{neurips_2024}

\usepackage[utf8]{inputenc} % allow utf-8 input
\usepackage[T1]{fontenc}    % use 8-bit T1 fonts
\usepackage{hyperref}       % hyperlinks
\usepackage{url}            % simple URL typesetting
\usepackage{booktabs}       % professional-quality tables
\usepackage{amsfonts}       % blackboard math symbols
\usepackage{nicefrac}       % compact symbols for 1/2, etc.
\usepackage{microtype}      % microtypography
\usepackage{xcolor}         % colors
\usepackage[title]{appendix}

\usepackage{enumitem}

\usepackage{graphicx}
\usepackage{algorithm}
\usepackage{algorithmic}
\usepackage{epstopdf,float,multirow,hyperref}

\usepackage[caption=false,font=footnotesize]{subfig}
\usepackage{amsmath}

\title{DisenGCD: A Meta Multigraph-assisted Disentangled Graph Learning Framework for  Cognitive Diagnosis}

\author{%
  Shangshang Yang$^{1,2}$ \And Mingyang Chen$^{1,3}$ \And
 Ziwen Wang$^{1}$~\And~Xiaoshan Yu$^{1}$~\And Panpan Zhang$^{1}$ \And Haiping Ma$^{1,3}$\thanks{Corresponding author.}~\And Xingyi Zhang$^{1,3}$ \\	 %\footnotemark[1]
    	$^1$Key Laboratory of Intelligent
Computing and Signal Processing of Ministry of Education, \\
Anhui University, Hefei, Anhui 230601,  P. R. China \\
$^2$ Anhui Province Key Laboratory of Intelligent Computing and Applications,\quad\\
$^3$Department of Information Materials and Intelligent Sensing Laboratory of Anhui Province\\
	\texttt{\{yangshang0308, wzw12sir, yxsleo, zppan55, xyzhanghust\}@gmail.com}\\ \texttt{{q22201127@stu.ahu.edu.cn}} \quad  \texttt{hpma@ahu.edu.cn}\\
}

\begin{document}

\maketitle

\begin{abstract}
 Existing graph learning-based cognitive diagnosis (CD) methods have made relatively good results, but their student, exercise, and concept representations are learned and exchanged in an implicit unified graph, which makes the interaction-agnostic exercise and concept representations be learned poorly, failing to provide high robustness against noise in students' interactions. Besides,  lower-order exercise latent representations obtained in shallow layers are not well explored when learning the student representation. 
To tackle the issues, this paper suggests a meta multigraph-assisted disentangled graph learning framework for CD (DisenGCD), which learns three types of representations on three disentangled graphs: student-exercise-concept interaction,  exercise-concept relation, and concept dependency graphs, respectively. 
Specifically,  the latter two graphs are first disentangled from the interaction graph. 
Then, the student representation is learned from the interaction graph by a devised meta multigraph learning module; multiple learnable propagation paths in this module  enable current student latent representation to access  lower-order exercise latent representations,
which can lead to  more effective nad robust student representations learned; 
the exercise and concept representations are learned on the relation and dependency graphs by graph attention modules. 
Finally, a novel diagnostic function is devised to handle three disentangled representations for prediction.  Experiments show better performance and robustness of DisenGCD than state-of-the-art CD methods and demonstrate the effectiveness of the disentangled learning framework and meta multigraph module.  
The source code is available at \textcolor{red}{\url{https://github.com/BIMK/Intelligent-Education/tree/main/DisenGCD}}.
\end{abstract}

\section{Introduction}
\begin{figure}[t]
   \centering
    \includegraphics[width=0.6\linewidth]{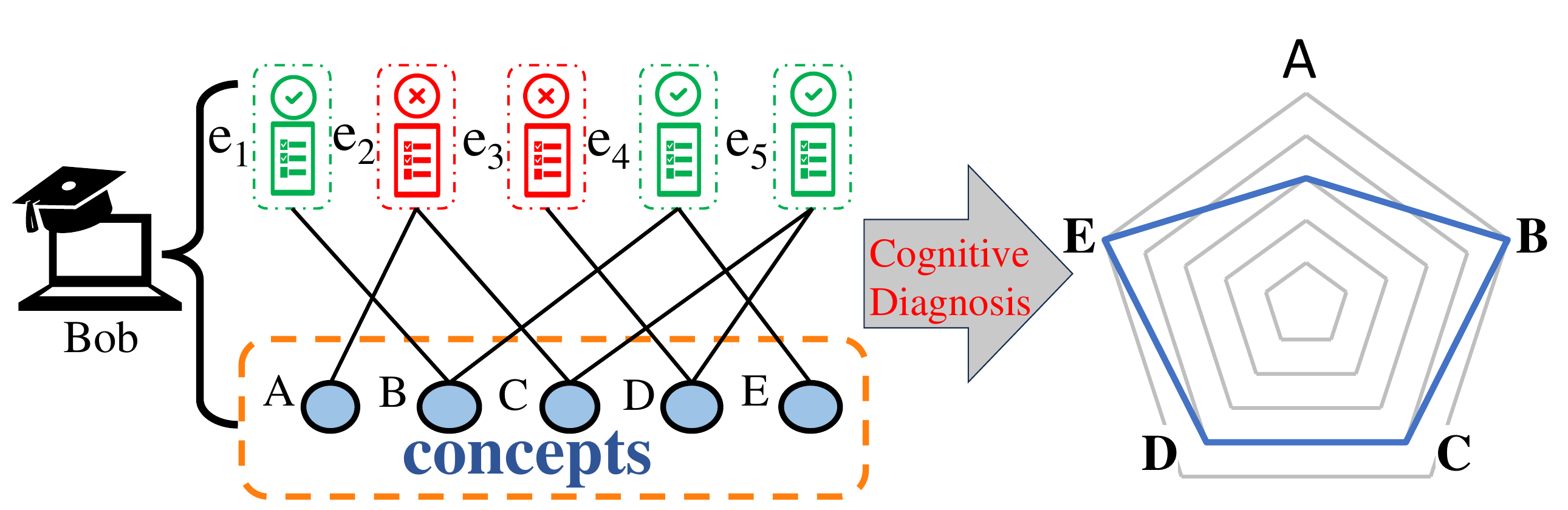}
      \caption{Cognitive diagnosis   based on a student's response records and  exercise-concept relations. }
      \label{introduction}
    \centering
\end{figure}

In the realm of intelligent education~\cite{ENAS-KT,wang2019mcne,yu2024rigl},
cognitive diagnosis~(CD) plays a crucial role in estimating students' mastery/proficiency on each knowledge concept~\cite{onlinecourses}, 
which mainly models the exercising process of students by predicting students' responses based on their response records/logs and the exercise-concept relations.
As shown in Figure~\ref{introduction}, student Bob completed five exercises $\{e_1,e_2,e_3,e_4,e_5\}$ and got corresponding responses,   
and his diagnosis result can be obtained through CD based on his records and the plotted relations between exercises and concepts. 
With students’ diagnosis results, many intelligent education tasks can be benefited, such as exercise assembly~\cite{yang2023cognitive}, course recommendation~\cite{courserecommendation,yin2024dataset}, student testing~\cite{dlcat}, and targeted training~\cite{beck2007difficulties}, and remedial instruction~\cite{liu2019exploiting}.

In recent years, many cognitive diagnosis models (CDMs) have been suggested based on various techniques, generally classified into two types.
The first type of CDMs is based on the theory of educational psychology, and the representatives include  DINA~\cite{dina},  (IRT)~\cite{irt}, and  MIRT~\cite{mirt}.
Considering that the simple diagnostic functions of this type of models fail to model complex student interactions, 
recently, artificial intelligence researchers have tried to employ neural networks (NNs) to enhance and invent novel CDMs. Therefore, the second type of CDMs can be further divided into two sub-types: focusing on inventing novel diagnostic functions and enhancing the input representations of CDMs, respectively.
 The representatives of the first sub-type of CDMs include the well-known NCD~\cite{ncd}, KSCD~\cite{kscd}, and so on~\cite{yang2023evolutionary,ma2024enhancing,liu2024automated,yang2024evolutionary,yang2024endowing}.
 For the second sub-type of CDMs, many NN techniques are employed, 
 including hierarchical attention NNs for ECD~\cite{ecd}, long short-term memory networks for DIRT~\cite{cheng2019dirt}, graph neural networks (GNNs) for  RCD~\cite{rcd}, SCD~\cite{scd}, and GraphCDM~\cite{gcdm}.

 Among existing CDMs, GNN-based ones exhibit significantly better performance than others, which is attributed to the rich information propagation and exchange brought by GNNs.
 These CDMs generally learn the student, exercise, and concept representations implicitly or explicitly on a unified graph~\cite{rcd,scd}, which makes three types of representations be exchanged and aggregated fully even if they are distant regarding relations, thus generating better representations and providing better performance.
 However,  learning exercise and concept representations should be student-interaction-agnostic~\cite{cdgk}, while the above CDMs' learning for exercise and concept representations is easily affected by students' interactions, and the representations will be learned poorly especially when there exists noise in students' interactions. 
 In short,  existing GNN-based CDMs fail to provide high robustness against student interaction noise.
 In addition,  these CDMs do not explore the use of lower-order exercise latent representations in shallow layers for learning the student representation due to the intrinsic defect of their  GNNs. The learned student representations could be more robust and not sensitive to student interaction data change.

Therefore, we propose a meta multigraph-assisted disentangled graph learning framework for CD (DisenGCD) to learn robust representations against interaction noise. The contributions include

    (1) The disentangled graph learning framework DisenGCD learns three types of representations on three disentangled graphs. Specifically, the student representation is learned on the student-exercise-concept interaction graph; the exercise and concept representations are learned on two disentangled graphs: the exercise-concept relation graph and the concept dependency graph. By doing so, the latter two learned representations are interaction-agnostic and robust against student interaction noise.
    
    (2) To make the best of lower-order exercise latent representations for learning the robust student representation, a modified meta multigraph module containing multiple learnable propagation paths is used, where propagation paths enable current student latent representation to access and use lower-order exercise latent representations.  
    The exercise and concept representations are learned through common graph attention networks (GAT) on relation and dependency graphs, respectively. Finally, a novel diagnostic function is devised to handle three learned representations for prediction.
    
    (3)  Extensive experiments show the superiority of the proposed DisenGCD to state-of-the-art (SOTA) models regarding performance and robustness, and the effectiveness of the disentangled graph learning framework and the devised meta multigraph module is validated.

\section{Related Work}
\textbf{Related Work on Cognitive Diagnosis}.
The above has given the classification of CDMs, and we will briefly introduce typical CDMs, especially  GNN-based ones. As representatives in educational psychology, IRT~\cite{irt} (MIRT~\cite{mirt}) utilizes single  (multiple) variable(s) to denote student's ability with logistic function for prediction. For NN-based representatives, NCD~\cite{ncd} and KSCD~\cite{kscd} fed student, exercise, and concept vectors to an IRT-like NN as the diagnostic function for prediction;  NAS-GCD~\cite{yang2023evolutionary} is similar to NCD, but its diagnostic function is automatically obtained.

For GNN-based CDMs, their focus is on obtaining enhanced representations.
For example, RCD~\cite{rcd} learns the student representation along the relation of adjacent exercise nodes, learns the exercise representation along the relation of adjacent student and concept nodes, and learns the concept representation along the relation of adjacent exercise and concept nodes. Since each node's information can be propagated to any node, learning regarding three relations can be seen as learning on an implicit unified graph; Similarly, SCD~\cite{scd} takes the same aggregation manner along the first two relations of RCD for learning student and exercise representations for contrastive learning to mitigate long-tail problems. Despite the success of GNN-based CDMs,  their learning manners of exercise and concept representations are not robust against student interaction noise, because the representations are learned together with student interactions in a unified graph,  where the noise will prevent the representations from being learned well. 

\textbf{Related Work on Meta Graph/Multigraph}.
As can be seen, existing GNN-based CDMs intuitively adopt classical GNNs (GAT \cite{GAT} or GCN\cite{lightgcn}) to learn three types of representations. 
 These CDMs update the student representation by only using the exercise latent representation in the previous layer yet not exploring the use of lower-order exercise latent representations.
 
 Compared to traditional GNNs, meta graph-based  GNNs can make the target type of nodes access lower-order latent representations of their adjacent nodes. 
 The meta graph is a directed acyclic graph built for a GNN, each node stores each layer's output (latent representations) of the GNN, and each edge between two nodes could be one  of   multiple types of propagation paths. The representatives  include DiffMG~\cite{diffmg} and  GEMS~\cite{gems}. 
The meta multigraph is the same as the meta graph, but its each edge could hold more than one type of propagation path, where the representative is PMMM~\cite{PMMM}. 
To make the best of low-order exercise latent representations for learning effective and robust student representations, this paper adopts the idea of meta multigraph and devises a modified meta multigraph learning module for the updating of the student interaction graph.    

\textbf{Related Work on Disentangling Graph Representation Learning}.
Learning potential representations of disentangling in complex graphs to achieve model robustness and interpretability has been a hot topic in recent years. Researchers have put forward many disentanglement approaches (e.g., DisenGCN~\cite{Ma2019DisentangledGC}, DisenHAN~\cite{Wang2020DisenHANDH}, DGCF~\cite{Wang2020DisentangledGC}, DCCF~\cite{DCCF}, DcRec~\cite{DcRec}, etc.) to address this challenge. For example, in DisenHAN~\cite{Wang2020DisenHANDH}, the authors utilized disentangled representation learning to account for the influence of each factor in an item. They achieved this by mapping the representation into different spatial dimensions and aggregating item information from various edge types within the graph neural network to extract features from different aspects; 
%In DCCF~\cite{DCCF}, the authors introduced global intent disentanglement into graph contrastive learning, extracting more fine-grained latent factors from self-supervised signals to enhance model robustness;
DcRec~\cite{DcRec} disentangles the network into a user-item domain and a user-user social domain, generating two views through data augmentation and ultimately obtaining a more robust representation via contrastive learning.

Despite many approaches suggested, they were primarily applied to bipartite graphs to learn different representations from different perspectives, for learning more comprehensive representations. While this paper aims to leverage disentanglement learning to mitigate the influence of the interaction noise in the interaction graph, and thus we proposed a meta multigraph-assisted disentangled graph cognitive diagnostic framework to learn three types of representations on three disentangled graphs. By doing so, the influence of the noise on exercise and concept learning can be well alleviated.

\section{Problem Formulation}
For easy understanding, two tables are created to describe all notations utilized in this paper including 
notations for disentangled graphs and notations for the meta multigraph, summarized in Table~\ref{tab:decopling} and Table~\ref{tab:meta multigraph}. Due to the page limit, the two tables are included in \textbf{Appendix}~\ref{sec:notations}.

\subsection{Disentangled Graph}
This paper only employs the student-exercise-concept interaction graph $\mathcal{G_I}$ for students' representation learning, and disentangles two graphs from $\mathcal{G_I}$ for the remaining two types of representation learning, which are the exercise-concept relation graph $\mathcal{G_R}$, and the concept dependency graph $\mathcal{G_D}$.

\noindent \textbf{Student-Exercise-Concept Interaction Graph.}
With  students' response records $\mathcal{R}$,  exercise-concept relation matrix, and 
   concept dependency matrix, the interaction graph $\mathcal{G_I}$ can be represented as $\mathcal{G_I}=\{\mathcal{V},\mathcal{E}\}$.   The node set $\mathcal{V} = S \cup E \cup C$ is the union of the student set $S$, exercise set $E$, and concept set $C$, 
 while the edge set $\mathcal{E} = \mathcal{R}_{se} \cup \mathcal{R}_{ec}   \cup \mathcal{R}_{cc}$ contains three types of relations:
 $rse_{ij} \in R_{se}$  represents the  student $s_i\in S$ answered exercise $e_j \in E$, 
  $rec_{jk} \in R_{ec}$  denotes the exercise $e_j$  contains concept $c_k \in C$, 
  and   $rcc_{km} \in R_{cc}$  denotes concept $c_k$ relies on concept $c_m$.

\noindent \textbf{Disentangled Relation Graph.}
 To avoid the impact of  students' interactions $ \mathcal{R}_{se}$ on exercise representation learning, 
 the   exercise-concept relation graph $\mathcal{G_R}$ is disentangled from $\mathcal{G_I}$, 
denoted as $\mathcal{G_R}= \mathcal{G_I}/ \{S, \mathcal{R}_{se} \}=  \{E \cup C,\mathcal{R}_{ec}   \cup \mathcal{R}_{cc}\}$.

 \noindent \textbf{Disentangled Dependency Graph.}
 Similarly,  a  concept dependency graph $\mathcal{G_D}$ is further  disentangled from $\mathcal{G_I}$ 
 to learn the concept representation without the influence of interactions $ \mathcal{R}_{se}$ and 
 exercises' relation $ \mathcal{R}_{ec}$, which is represented by 
 $\mathcal{G_D}= \mathcal{G_R}/ \{E, \mathcal{R}_{ec} \} = \{ C,\mathcal{R}_{cc}\}$.

 \subsection{Problem Statement}
For the cognitive diagnosis task in an intelligent education online platform, 
there are usually three sets of items: 
a set of $N$ students $S = \{ s_1, s_2, \ldots, s_N \}$,
a set of $M$ exercises $E = \{ e_1, e_2, \ldots, e_M \}$,
and a set of $K$ knowledge concepts (concept for short) $ C = \{ c_1, c_2, \ldots, c_K \}$. 
Besides, there commonly exists two matrices:  exercise-concept relation  matrix $Q=(Q_{jk}\in\{0,1\})^{M\times K}$ (Q-matrix) and  concept dependency matrix $D=(Q_{km}\in\{0,1\})^{K\times K}$, to show the relationship of exercises to concepts and concepts to concepts, respectively. $Q_{jk}=1$ denotes the concept $c_k$ is not included in the exercise $e_j$  and  $Q_{jk}=0$ otherwise;
similarly, $D_{km}=1$ denotes  concept $c_k$ relies on  concept $c_m$   and  $D_{km}=0$ otherwise.
All students' exercising reponse logs  are denoted by $\mathcal{R}=\{(s_i,e_j,r_{ij})|s_i\in S, e_j\in E, r_{ij}\in\{0,1\}\}$, where $r_{ij}$ refers to the response/answer of student $s_i$ on exercise $e_j$. $r_{ij}=1$ means the answer is correct and $r_{ij}=0$ otherwise. 

 The CD based on disentangled graphs is defined as follows: \textbf{Given}: students' response logs $\mathcal{R}$ and three disentangled graphs: interaction graph $\mathcal{G_I}$, relation graph $\mathcal{G_R}$, and dependency graph $\mathcal{G_D}$;
\textbf{Goal}: revealing students'  proficiency on concepts by predicting students' responses through  NNs.

\begin{figure*}[t]
    \centering
    \includegraphics[width=1.\linewidth]{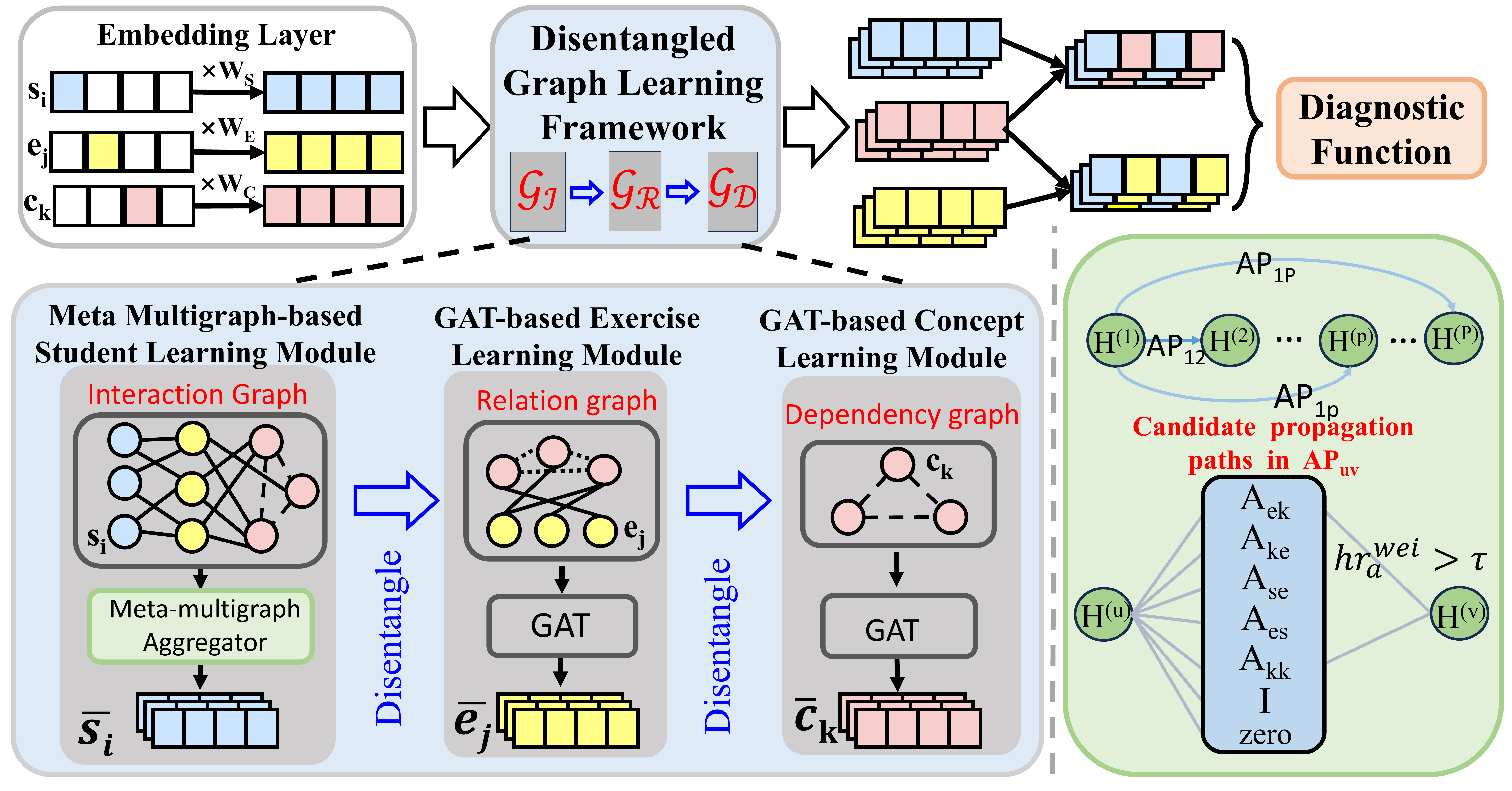}
    \caption{Overview of the proposed  DisenGCD: the disentangled graph learning framework is composed of three learning modules:  two GAT-based learning modules and a meta multigraph-based learning module, where the latter's details are shown in the green part. }
    \label{figure2}
\end{figure*}
\section{Method}

For better understanding, Figure~\ref{figure2} presents the overall architecture of the proposed DisenGCD, where three learning modules are used for learning three types of representations on three disentangled graphs and the diagnostic function makes the final prediction based on the learned representations.
Specifically, these three learning modules are:  (1) a meta multigraph-based student learning module, (2) 
a GAT-based exercise learning module, and (3) a GAT-based concept learning module. 
In each learning module, the corresponding graph's embedding is randomly first initialized.
Then, the meta multigraph-based student learning module is employed to learn the student  representation $\overline{\mathbf{S}_i}$ based on interaction graph  $\mathcal{G_I}$;
the GAT-based exercise learning module is used to learn the exercise  representation $\overline{\mathbf{E}_j}$ based on relation graph  $\mathcal{G_R}$;
while the concept representation $\overline{\mathbf{C}_k}$ is learned by the GAT-based concept learning module on dependency graph $\mathcal{G_D}$ or the naive embedding if $\mathcal{G_D}$ is unavailable.
Finally, a devised novel diagnostic function receives three learned representations $\overline{\mathbf{S}_i}$, $\overline{\mathbf{E}_j}$, and $\overline{\mathbf{C}_k}$ to get the prediction $\hat{r_{ij}}$ of student $s_i$ on exercise $e_j$.

Note that the first module employs a modified meta multigraph aggregator to learn student representations.
Compared to traditional graph learning, 
the module contains multiple learnable propagation paths, which enable the student latent representation to be learned currently to access and use lower-order exercise latent representations, leading to a more effective and robust student representation.

\subsection{The Meta Multigraph-based Student Learning Module}

%H^(i)属于节点类型集T（即学生、练习、概念三类），H^(0)和H^(n)是元多图的输入和输出节点，且输出的H^(n)作为更新后的表征进行输出。在元多图的搜索过程当中，当确定超节点中的边类型后，则表示该边的初始节点则沿着这条边聚合目标节点的表征；例如：A(0,1)=EK,则表示在该阶段练习节点对概念表征进行聚合。

%\subsection{The Meta-Multigraph Module}

Based on interaction graph $\mathcal{G_I}$, this module is responsible for: learning the optimal meta multigraph structure in $\mathcal{G_I}$, i.e.,  optimal multiple propagation paths, and updating student nodes' representations based on the learned meta multigraph structure to get the representation $\overline{\mathbf{S_i}}\in \mathbb{R}^{1\times d}$.

To start with, all nodes (including students, exercises, and concepts) need to be mapped into a $d$-dimensional  hidden space. The student embedding $\mathbf{s}_{i}\in \mathbb{R}^{1\times d}$ for student $s_i$, 
 exercise  embedding $\mathbf{e}_{j}^{I}\in \mathbb{R}^{1\times d}$ for exercise $e_j$, 
and  concept embedding $\mathbf{c}_{k}^I\in \mathbb{R}^{1\times d}$ for concept $c_k$ can be obtained by
\begin{equation}\label{eq:embedding}
\mathbf{s}_{i} = \mathbf{x}_{i}^S\times W_S^I, \mathbf{e}_{j}^I = \mathbf{x}_{j}^{E}\times W_E^I, \mathbf{c}_{k}^I = \mathbf{x}_{k}^{C}\times W_C^I,  W_S^I \in \mathbb{R}^{N\times d}, W_E^I \in \mathbb{R}^{M\times d},  W_C^I \in \mathbb{R}^{K\times d}.
\end{equation}
 $\mathbf{x}_{i}^S\in \{0,1\}^{1\times N}$, $\mathbf{x}_{j}^E\in \{0,1\}^{1\times M}$, and $\mathbf{x}_{k}^{C}\in \{0,1\}^{1\times K}$ are three one-hot vectors for   student $s_i$, exercise $e_j$, concept $c_k$, respecitvely.
$W_S^I$, $W_E^I$, and $W_C^I$ are three learnable parameter matrices.

\subsubsection{The Meta Multigraph Aggregator}
To learn more useful representations from the above initial embedding,  we construct a meta multigraph $\mathcal{MG}=\{\mathbf{H}, \mathcal{AP}\}$ to specify a set of propagation paths for all nodes' aggregation in graph learning.

As shown in Figure~\ref{figure2}, the meta multigraph is actually a directed acyclic graph containing $P$ hyper-nodes: $\mathbf{H} = \{\mathbf{H}^{(1)}, \dots, \mathbf{H}^{(p)}, \dots, \mathbf{H}^{(P)} \}$. $\mathbf{H}^{(p)}=\{\mathbf{s}^p\in \mathbb{R}^{N\times d}, \mathbf{e}^p \in \mathbb{R}^{M\times d}, \mathbf{c}^p\in \mathbb{R}^{K\times d}\}$ is a set of latent representations stored in $p$-th hyper-node for all students, exercises, and concepts,
 whose $i$-th row $\mathbf{s}_i^p\in \mathbb{R}^{1\times d}$, $j$-th row $\mathbf{e}_j^p\in \mathbb{R}^{1\times d}$, and $k$-th row $\mathbf{c}_k^p\in \mathbb{R}^{1\times d}$ are the $p$-th latent representations of student $s_i$, exercise $e_j$, and concept $c_k$,  respectively. Especially when $p$=1, $\mathbf{s}_i^p$,  $\mathbf{e}_j^p$, and $\mathbf{c}_k^p$  equal  initial embedding $\mathbf{s}_i$, $\mathbf{e}_j^I$, and $\mathbf{c}_k^I$.

Each edge $AP_{uv}\in \mathcal{AP}$ between two hyper-nodes $H^{(u)}$ and $H^{(v)}$ contains multiple types ($|HR|$) of propagation paths, which can be denoted as $AP_{uv}=\{(hr_a\in \mathcal{HR}, hr_a^{wei}\in [0,1])| 1 \leq a \leq |HR| \}$.
Here  $hr_a$ is $a$-th  type of propagation path  in candidate path set  $HR$, and 
$hr_a^{wei}$ represents the weight  of $hr_a$ . 
This paper adopts $|HR|$=7 types of propagation paths as the candidate paths for  $HR$: 1) the students to exercises path $A_{se}$, 2) the exercises to students  path $A_{es}$, 3) the exercises to concepts path $A_{ek}$, 4) the concepts to exercises path $A_{ke}$, 5) the concepts to concepts path $A_{kk}$, 6)  the identity path $I$, 7)  the zero path $zero$. 

Here, path $A_{se}$ means updating the exercise latent representation along the propagation path of student nodes to exercise nodes, and other paths hold similar meanings.
As a result, $HR=\{A_{se}, A_{es}, A_{ek}, A_{ke}, A_{k\hat{k}}, I, zero\}$, and 
the edge set $\mathcal{AP}$ in $\mathcal{MG}$  can be denoted by $\mathcal{AP}=\{AP_{uv} |1\leq u < v \leq P\}$, containing $|HR|*[P*(P-1)/2]$ propagation paths.
With  above propagation paths, $\mathbf{H}^{(p)}$ can be updated by
\begin{equation}
  \mathbf{H}^{(p)}=\sum\{  f(\mathcal{AP}_{up}, \mathbf{H}^{(u)})|\ 1\leq u <p\},
  \label{equation4}
\end{equation}
where $f(\cdot)$ is the aggregation function of GCN.
$f(\mathcal{AP}_{up}, \mathbf{H}^{(u)})$ refers to 
that obtaining the latent representations ($\mathbf{s}_i^p, \mathbf{e}_j^p, \mathbf{c}_k^p$)  in hyper-node 
$\mathbf{H}^{(p)}$ based on all latent representations stored in previous hyper-nodes (i.e., $\mathbf{H}^{(1)}$ to $\mathbf{H}^{(p-1)}$).
It can be seen that the updating process of $\mathbf{H}^{(p)}$ replies on latent representations in multiple hyper-nodes and multiple propagation paths, which compose \textbf{the modified meta multigraph aggregator} together.

For better understanding, here is an example: when $p$=3,   $AP_{13}=\{zero\}$ and $AP_{23}=\{A_{es},  I\}$, the updating of three  latent representations $\mathbf{s}_i^3$, $\mathbf{e}_j^3$, and $\mathbf{c}_k^3  $ is denoted as
 \begin{equation}
 \begin{aligned}
 &AP_{13}:\left\{ \mathbf{s}_i^3(13) \textrm{=} 0*\mathbf{s}_i^1, \mathbf{e}_j^3(13)  \textrm{=} 0*\mathbf{e}_j^1 , \mathbf{c}_k^3(13)  \textrm{=} 0*\mathbf{c}_k^1\right.\\
 &AP_{23}:\ \left\{ 
  \begin{aligned}
\mathbf{s}_i^3(23) = Up(\mathbf{s}_i^2, \sum_{j \in N_{s_i}} Mess(\mathbf{s}_i^2, e_j^{2}))\\ 
 \mathbf{e}_j^3(23) =  \mathbf{e}_j^{2},  \mathbf{c}_k^3(23) =  \mathbf{c}_k^{2}\\ 
  \end{aligned}\right.\\
 & \mathbf{s}_i^3 = \mathbf{s}_i^3(13) +\mathbf{s}_i^3(23),\  
  \mathbf{e}_j^3 =  \mathbf{e}_j^3(13) +\mathbf{e}_j^3(23),\  \mathbf{c}_k^3  =   \mathbf{c}_k^3(13) +   \mathbf{c}_k^3(23) \\
  \end{aligned}.
 \end{equation}
 The first equation is to update  three latent representations according to $AP_{13}$, while the second is to get the updating based on $AP_{23}$. 
 The final latent representations are obtained by summing up both updated ones. Here $Up(\cdot)$ and $Mess(\cdot)$ are common update and message-passing functions.

\subsubsection{Routing Strategy}
To find suitable propagation paths, threshold $\tau^{(u, v)}$ is created  for each pair of  hyper-nodes $(u,v)$:
\begin{equation}
\begin{aligned}
\tau^{(u, v)}=\lambda \cdot \max ( Softmax(AP_{uv}))+(1-\lambda) \cdot  \min (Softmax(AP_{uv})), \ \lambda\ \textrm{is predefined}\\
\end{aligned}.
\end{equation}
 $Softmax(AP_{uv}))$ normalizes  weights of each type of propagation path regarding their $hr_a^{wei}$ values. 
By doing so,  the propagation paths of each pair of hyper-nodes will remain if their $hr_a^{wei}$ values are greater than the corresponding threshold.
Thus the  updating process of $\mathbf{H}^{(p)}$ can be rewritten as 
\begin{equation}
\begin{aligned}
\mathbf{H}^{(p)}&=\sum  \{f(\hat{AP_{up}} , \boldsymbol{H}^{(u)})||\ 1\leq u <p\}\\
\hat{AP_{up}} &= \{ (hr_a, hr_a^{wei})| hr_a^{wei} \geq \tau^{(u,p)}, \forall hr_a \in AP_{up} \}\\
\end{aligned}.
  \label{eq: rewr}
\end{equation}
Finally, the learned student representation $\mathbf{s}^P$ in $\mathbf{H}^{P}$ are used for the  diagnosis, i.e, $\mathbf{s}_i^P$ is used as  $\overline{\mathbf{S}_i}$.

\subsection{The GAT-based Exercise Learning Module and  GAT-based Concept Learning Module}

\textbf{GAT-based Exercise Learning Module}. This module is responsible for learning the exercise  representation $\overline{\mathbf{E_j}} \in \mathbb{R}^{1\times d}$ on the relation graph $\mathcal{G_R}$ via a $L$-layer GAT network~\cite{GAT}.
Firstly, the  embedding of exercises and concepts in $\mathcal{G_R}$  is obtained in the  manner same as Eq.(\ref{eq:embedding}) through two learnable matrices $ W_E^R\in \mathbb{R}^{M\times d}$ and $ W_C^R\in \mathbb{R}^{K\times d}$, i.e, $\mathbf{e}_{j}^R = \mathbf{x}_{j}^{E}\times W_E^R, \mathbf{c}_{k}^R = \mathbf{x}_{k}^{C}\times W_C^R$.
%where $\mathbf{e}_{j}^R\in \mathbb{R}^{1\times d} $ and $\mathbf{c}_{k}^R\in \mathbb{R}^{1\times d}$ are the embedding for exercise $e_j$ and   concept $c_k$. 
%$ W_E^R\in \mathbb{R}^{M\times d}$ and $ W_C^R\in \mathbb{R}^{K\times d}$ are two learnable matrices in this module.

Afterward, the GAT neural network is applied to aggregate neighbor information to learn the exercise representation. The aggregation process of $l$-th layer ($1\leq l\leq L$) can be represented as 
\begin{equation}
\small
    \begin{aligned}
     &\mathbf{e}_{j}^{R(l)} = \textstyle \sum_{k \in N_{e_{j}}} \alpha_{j (k)}^{R(l)} \mathbf{c}_k^{R(l-1)}+\mathbf{e}_j^{R(l-1)},\\ 
     &\mathbf{c}_{k}^{R(l)} = 
      \textstyle \sum_{j \in N_{c_{k}}^{ec}} \alpha_{k(j) }^{R(l)} \mathbf{e}_j^{R(l-1)}+
      \textstyle \sum_{m \in N_{c_{k}}^{cc}} \alpha_{\hat{k}(m) }^{R(l)} \mathbf{c}_m^{R(l-1)}+
      \mathbf{c}_k^{R(l-1)}\\
    \end{aligned}.
\end{equation}
 The first equation is to aggregate the information of exercise $e_j$'s neighbors $N_{e_{j}}$  to get its $l$-th layer's latent representation $\mathbf{e}_{j}^{R(l)}\in \mathbb{R}^{1\times d}$, while the second is to update the $l$-th layer's concept latent representation  $\mathbf{c}_{k}^{R(l)}\in \mathbb{R}^{1\times d}$ from its exercise neighbors $N_{c_{k}}^{ec}$ and concept neighbors $N_{c_{k}}^{cc}$.  $\alpha_j^{R(l)}$, $\alpha_{k}^{R(l)}$, and $\alpha_{\hat{k}}^{R(l)}$ are  the $l$-th layer's  attention matrices for exercise $e_j$'s concept neighbors,  concept $c_k$'s exercise neighbors, and $c_k$'s concept neighbors. 
They can be obtained in the same manner, and the $k$-th row of  $\alpha_j^{R(l)}$ can be obtained by
\begin{equation}\label{eq:atten}
  \alpha_{j (k)}^{R(l)} = {\rm Softmax}( F_{ec}([\mathbf{e}_{j}^{R(l-1)}, \mathbf{c}_{k}^{R(l-1)}])), \forall k\in N_{e_j}.
\end{equation}
   $[ \cdot ]$ is the concatenation  and $F_{ec}(\cdot)$ is a fully connected (FC) layer mapping  $2*d$ vectors to  scalars. 

 The latent representation $\mathbf{e}_j^{R(0)}$ and  $\mathbf{c}_k^{R(0)}$ refer to $\mathbf{e}_j^{R}$ and  $\mathbf{c}_k^{R}$, and the $L$-th layer output $\mathbf{e}_{j}^{R(L)} $ is used as $\overline{\mathbf{E}_j}$.
By disentangling the interaction data, the learned exercise  representation $\overline{\mathbf{E}_j}$ will be more robust against interaction noise.

%\subsection{The GAT-based Concept Learning Module}
\textbf{GAT-based Concept Learning Module}. Similarly, this module obtains the concept  representation $\overline{\mathbf{C}_k}\in \mathbb{R}^{1\times d}$ by applying a $L$-layer GAT network to the dependency graph $\mathcal{G_D}$.
After obtaining the initial embedding $\mathbf{c}_k^D = \mathbf{x}_k^C\times W_C^D$ by a learnable matrix $ W_C^D\in\mathbb{R}^{K\times d}$, 
this module updates the $l$-th layer's latent representation $\mathbf{c}_{k}^{D(l)}$  by 
$\mathbf{c}_{k}^{D(l)} =  \textstyle \sum_{m \in N_{c_{k}}^{cc}} \alpha_{\hat{k}(m) }^{D(l)} \mathbf{c}_m^{D(l-1)}+ \mathbf{c}_k^{D(l-1)}$.
$\alpha_{\hat{k}}^{D(l)}$ denotes the attention matrix, which can be computed as same as Eq.(\ref{eq:atten}).

%which is used to aggregate the information of concept $c_k$'s neighbors to get $\mathbf{c}_{k}^{D(l)}$. 

Here $\mathbf{c}_{k}^{D(0)}$ is equal to $\mathbf{c}_k^D$, and the $L$-th layer output $\mathbf{c}_{k}^{D(L)} $ is used as $\overline{\mathbf{C}_k}$. If graph $\mathcal{G_D}$ is unavailable, this module will directly take the initial embedding $\mathbf{c}_k^D$ as  $\overline{\mathbf{C}_k}$.
By further disentangling, the learned representation $\overline{\mathbf{C}_k}$ may be robust against noise in student interactions to some extent.

\subsection{The Diagnosis Module}
To effectively handle  the obtained three types of  representations, 
a novel diagnostic function is proposed to predict the response $\hat{r_{ij}}$ of student $s_i$ got on exercise $e_j$ as follows:
\begin{equation}
    \begin{aligned}
     \mathbf{h}_{simi} = \sigma(F_{simi}(\mathbf{h}_{si}\cdot\mathbf{h}_{ej}  )), \mathbf{h}_{si} &= F_{si}(\overline{\mathbf{S}_i}+ \overline{\mathbf{C}_k}), \mathbf{h}_{ej} = F_{ej}(\overline{\mathbf{E}_j}+ \overline{\mathbf{C}_k})\\
    \hat{r_{ij}} &= (\sum Q_k\cdot \mathbf{h}_{simi})/\sum Q_k \\
    \end{aligned},
\end{equation}
where $F_{si}(\cdot)$,  $F_{ej}(\cdot)$, and  $F_{simi}(\cdot)$ are three FC layers mapping a $d$-dimensional vector to another one, and $Q_k$ is a binary vector in the $k$-th row of Q-matrix.

Here $\mathbf{h}_{si}\in \mathbb{R}^{1\times d}$  can be seen as  the  student's mastery of each knowledge concept;
while the obtaining of  $ \mathbf{h}_{ej} \in \mathbb{R}^{1\times d}$  aims to get the exercise difficulty of each concept;
$ \mathbf{h}_{simi}$ is to measure the similarity between $\mathbf{h}_{si}$ and $\mathbf{h}_{ej}$ via a dot-product followed by an FC layer and a Sigmoid function $\sigma(\cdot)$~\cite{yang2022accelerating}, where a higher similarity value in each bit represents a higher mastery on each concept, further indicating a higher probability of answering the related exercises;
the last equation follows the idea of NCD to compute the overall mastery averaged over all concepts contained in exercise $e_j$.

We can see that the proposed diagnostic function has as high interpretability as  NCD, IRT, and MIRT.

\textbf{Model Optimization.} 
With the above modules,  the proposed DisenGCD are trained by solving the following bilevel optimization problem through Adam~\cite{yang2024hybrid}: 
\begin{equation}
\begin{aligned}
\min _{\boldsymbol{\alpha}} \mathcal{L}_{ val }(D_{val}|\boldsymbol{\omega}^{*}(\boldsymbol{\alpha}), \boldsymbol{\alpha}),\ \text { s.t. } \boldsymbol{\omega}^{*} (\boldsymbol{\alpha})=\operatorname{argmin}_{\boldsymbol{\omega}} \mathcal{L}_{ train }(D_{train}|\boldsymbol{\omega}, \boldsymbol{\alpha}),
\end{aligned}
\end{equation}
where \(\mathcal{L}_{val}(\cdot)\) and \(\mathcal{L}_{train}(\cdot)\) denote the loss on validation dataset $D_{val}$   and training dataset  $D_{train}$.
\(\boldsymbol{\omega}\) denotes all  model parameters, and $\boldsymbol{\alpha}$ denotes the weights of learnable propagation paths $\mathcal{AP}$ in meta multigraph $MG$.  The   cross-entropy  loss~\cite{yang2021gradient} is used for $ \mathcal{L}_{val}$ and $\mathcal{L}_{train}$.
%:
%\begin{equation}
%\mathcal{L}_(data)=-\sum_{r_{ij}\in data}(r_{ij}\log \hat{r_{ij}}+(1\textrm{-}r_{ij})\log(1\textrm{-}\hat{r_{ij}})).
%\end{equation}

%%%%%%%%%%%%%%%%%%%%%%%%%%%%%%%%%%%%%%%%%%%%%%%%%%%%%%%%%%%%

\section{Experiments}
This section  answers the following questions: $\mathbf{RQ1}$: How about the performance of  DisenGCD compared to SOTA CDMs? $\mathbf{RQ2}$: How about the DisenGCD's robustness against noise and  the disentangled learning framework's effectiveness in DisenGCD? $\mathbf{RQ3}$: How about the effectiveness of the devised meta multigraph learning module? $\mathbf{RQ4}$: How does the learned meta multigraph on  target datasets look like, 
    and how about their generalization on other datasets? In addition, \textbf{more experiments to validate the proposed DisenGCD's effectiveness are in the Appendix.}

\subsection{Experimental Settings}
\begin{table}[h]
\renewcommand{\arraystretch}{1.}
\small
    \centering
        \caption{Statistics of three datasets: ASSISTments, Math, and SLP.}

          \setlength{\tabcolsep}{.5mm}{
    \begin{tabular}{lccccc}
         \hline
           Datasets & Students & Exercises  & Concepts & Logs  & Avg logs per student \\
         \hline
         \# ASSISTments & 4,163 & 17,746 & 123 & 278,868
         & 67 \\
         \# Math / SLP & 1,967 / 1,499 & 1,686 / 907 & 61 / 33 & 118,348 / 57,244 & 60 / 38 \\
       %  \# SLP  & 1,499 & 907 & 33 & 57,244 & 38 \\
         \hline
    \end{tabular}
    }

    \label{tab:Table1}
\end{table}
\iffalse \begin{table}[h]
    \centering
    \caption{Statistics of three datasets}
    \label{tab:Table1}
    \begin{tabular}{llll}
        \hline
          Statistics & ASSISTment & Math & SLP\\
        \hline
        \# Students & 4,163  &  1,976 & 1499 \\
        \# Exercises & 17,746 & 1,686 & 907 \\
        \# Knowledge concepts & 123 & 61 & 33\\
        \# Response logs & 278,868 & 118,348 & 57,244 \\
        \# Avg logs per student & 67 & 60 & 38 \\
         \hline
        \end{tabular}
\end{table}
\fi
\textbf{Datasets.}\quad
To verify the DisenGCD's effectiveness, we conducted experiments on two public datasets ASSISTments~\cite{feng2009addressing} and SLP~\cite{lu2021slp},  and one private dataset Math, 
whose statistics are in Table~\ref{tab:Table1}. Note that ASSISTments and Math~\cite{ma2024enhancing} were used for most experiments to answer \textbf{RQ1} to \textbf{RQ3}, 
and SLP  was used to answer \textbf{RQ4}.  
\textbf{More details of these datasets are presented in} \textbf{Appendix~\ref{sec:data_details}}.

\noindent \textbf{Comparison CDMs and Metrics.}\quad 
To verify the effectiveness of  DisenGCD, five SOTA CDMs were compared, including traditional CDMs DINA and MIRT, NN-based CDMs NCD and ECD, and GNN-based RCD. 
Here SCD is not compared due to the failed run of its provided source code
 Three metrics were adopted to measure the performance of  all CDMs~\cite{yang2023designing}, including \textit{area under the cure }(AUC)~\cite{sun2024interpretable}, \textit{accuracy} (ACC)~\cite{sun2023adversarial}, and \textit{root mean square error} (RMSE)~\cite{yin2024entropy}.

\noindent \textbf{Parameter Settings.}
For the DisenGCD model, its dimension $d$ was set to the number of concepts,
$P$ in the meta multigraph, the number of layers $L$ in GAT,  and  $\lambda$ were set to 5, 2, and 0.8.
For its training, the learning rate and batch size were 1e-4 and 256.
For comprehensive comparisons~\cite{ma2024hd}, four splitting ratios (40\%/10\%/50\%, 50\%/10\%/40\%, 60\%/10\%/30\%, and 70\%/10\%/20\%) were adopted to get training, validation, and testing datasets. 
All compared CDMs followed the settings in their original papers,
 %DisenGCD was implemented by Python and PyTorch, 
 and all experiments were executed on an NVIDIA RTX4090 GPU.
 
\begin{table*}[t]
  \centering
  \caption{Performance comparison between DisenGCD and five CDM in terms of AUC, ACC, and RMSE values, obtained on ASSISTments and Math. Four dataset-splitting ratios were adopted, and the best result of each column on one dataset was highlighted. }
  \small
    \renewcommand{\arraystretch}{1.2}
  \setlength{\tabcolsep}{.2mm}{
     \begin{tabular}{c|c|ccc|ccc|ccc|ccc}
    \toprule
    Datatset & Ratio & \multicolumn{3}{c|}{40\%/10\%/50\%} & \multicolumn{3}{c|}{ 50\%/10\%/40\%} & \multicolumn{3}{c|}{ 60\%/10\%/30\%} & \multicolumn{3}{c}{70\%/10\%/20\%} \\
    \midrule
     & Method & ACC$\uparrow$   & RMSE$\downarrow$  & AUC$\uparrow$      &ACC$\uparrow$   & RMSE$\downarrow$  & AUC$\uparrow$&ACC$\uparrow$   & RMSE$\downarrow$  & AUC$\uparrow$& ACC$\uparrow$   & RMSE$\downarrow$  & AUC$\uparrow$\\
\cmidrule{2-14}          & DINA  & 0.6388 & 0.4931 & 0.6874 & 0.6503 & 0.4862 & 0.4978 & 0.6573 & 0.4820 & 0.7071 & 0.6623 & 0.4787 & 0.7126 \\
          & MIRT  & 0.6954 & 0.4740 & 0.7254 & 0.7015 & 0.4689 & 0.7358 & 0.7096 & 0.4624 & 0.7469 & 0.7110 & 0.4617 & 0.7514 \\
        ASSIST  & NCD   & 0.7070 & 0.4443 & 0.7374 & 0.7142 & 0.4370 & 0.7423 & 0.7237 & 0.4365 & 0.7552 & 0.7285  & 0.4298 & 0.7603 \\
         ments & ECD   & 0.7154 & 0.4373 & 0.7362 & 0.7130 & 0.4373 & 0.7432 & 0.7274 & 0.4329 & 0.7543 & 0.7297 & 0.4296 & 0.7599 \\
          & RCD   & 0.7232 & 0.4311 & 0.7546 & 0.7253 & 0.4285 & 0.7605 & 0.7291 & 0.4262 & 0.7663 & 0.7296 & 0.4245 & 0.7687 \\
          & DisenGCD & \textbf{0.7276} & \textbf{0.4255} & \textbf{0.7635} & \textbf{0.7287} & \textbf{0.4238} & \textbf{0.7677} & \textbf{0.7335} & \textbf{0.4219} & \textbf{0.7723} & \textbf{0.7334} & \textbf{0.4209} & \textbf{0.7746} \\
    \midrule
    \multirow{6}[2]{*}{Math} & DINA  & 0.6691 & 0.4715 & 0.7117 & 0.6745 & 0.4674 & 0.7199 & 0.6813 & 0.4633 & 0.7222 & 0.6812 & 0.4635 & 0.7231 \\
          & MIRT  & 0.7229 & 0.4335 & 0.7427 & 0.7227 & 0.4299 & 0.7497 & 0.7279 & 0.4291 & 0.7479 & 0.7340 & 0.4256 & 0.7542 \\
          & NCD   & 0.7394 & 0.4157 & 0.7604 & 0.7424 & 0.4119 & 0.7660 & 0.7418 & 0.4109 & 0.7706 & 0.7447 & 0.4084 & 0.7756 \\
          & ECD   & 0.7335 & 0.4154 & 0.7615 & 0.7424 & 0.413 & 0.7657 & 0.7434 & 0.4114 & 0.7693 & 0.7484 & 0.4087 & 0.7761 \\
          & RCD   & 0.7446 & 0.4100  & 0.7724 & 0.7489 & 0.4074 & 0.7751 & 0.7501 & 0.4078 & 0.7806 & 0.7534 & 0.4034 & 0.7866 \\
          & DisenGCD & \textbf{0.7479} & \textbf{0.4076} & \textbf{0.7802} & \textbf{0.7513} & \textbf{0.4052} & \textbf{0.7832} & \textbf{0.7527} & \textbf{0.4039} & \textbf{0.7867} & \textbf{0.7582} & \textbf{0.4004} & \textbf{0.7932} \\
    \bottomrule
    \end{tabular}%
    
    }
  \label{tab:main_experiment}%
\end{table*}%
\subsection{Overall Performance Comparison~(\textbf{RQ1})}\label{sec:RQ1}
To address \textbf{RQ1}, the DisenGCD was compared with  DINA, MIRT, NCD, ECD, and RCD  on ASSISTments and Math datasets. 
Table~\ref{tab:main_experiment} summarizes their performance in terms of AUC, ACC, and RMSE  obtained under four dataset-splitting settings, where
 the best result of each column on one dataset was highlighted in bold. Besides, Table~\ref{tab:sotaCDM1} in the \textbf{Appendix} compares DisenGCD with three SOTA CDMs, including SCD~\cite{scd}, KSCD~\cite{kscd}, and KaNCD~\cite{wang2022neuralcd}.
 
 We can  observe from the results in both tables that, 
 DisenGCD holds better performance than all compared CDMs. 
 Besides, we can also obtain two observations  from Table~\ref{tab:main_experiment}:
 Firstly, GNN-based CDMs (RCD and DisenGCD) hold significantly better than NN-based CDMs (NCD and ECD),  indicating the importance of learning representations through graphs, 
 and  DisenGCD outperforming RCD validates the superiority of DisenGCD's graph learning manner.
Secondly,  as the ratio changes, the performance change of DisenGCD and RCD is much smaller than NCD and ECD, which signifies  DisenGCD and RCD  are more robust to different-sparsity data.
To  validate its sparsity superiority, \textbf{more experiments} are summarized in \textbf{Appendix \ref{sec:exp_sparse}}.

\begin{table}[!t]
  \caption{Performance of DisenGCD,  RCD, and its four variants on ASSISTments dataset.}
  \label{ablation experiment}
\small
  \renewcommand{\arraystretch}{1.1}
  \centering
  \begin{tabular}{c|cccccc}
  \toprule
 \multirow{2}[2]{*}{Metric} & \multirow{2}[2]{*}{RCD} & \multirow{2}[2]{*}{\textit{DisenGCD(I)}} & \textit{DisenGCD} &\textit{DisenGCD} & \textit{DisenGCD} & \multirow{2}[2]{*}{DisenGCD} \\
  & & &  \textit{(Is+Rec)}& \textit{(Ise+Rc)}& \textit{(Isc+Re)} & \\
  \hline
  ACC$\uparrow$     & 0.7291  &0.7331 &0.7321 &0.7301 &0.7333 &\textbf{0.7335}\\
  RMSE$\downarrow$  &0.4262   &0.4235 &0.4259 &0.4235 &0.4231 &\textbf{0.4219}\\
  AUC$\uparrow$     &  0.7663 &0.7678 &0.7701 &0.7678 &0.7685 & \textbf{0.7723}\\
\bottomrule
  \end{tabular}

\iffalse
  \begin{tabular}{c|ccc}
        %\hline
        %& \multicolumn{3}{c|}{ASSISTments} \\ % & \multicolumn{3}{c}{Math}  \\
        \toprule
        Method & ACC$\uparrow$   & RMSE$\downarrow$  & AUC$\uparrow$  \\ %&ACC$\uparrow$   & RMSE$\downarrow$  & AUC$\uparrow$  \\
        \hline
        RCD & 0.7291 & 0.4262 & 0.7663  \\ %& 0.7501 & 0.4078 & 0.7806 \\
         \textit{DisenGCD(I)} & 0.7331 & 0.4235 & 0.7678  \\ %& 0.7510 & 0.4062 & 0.7829 

                          \textit{DisenGCD(Is+Rec)} & 0.7321 & 0.4259 &
        0.7701  \\ %& 0.7501 & 0.4055 & 0.7838
        
        \textit{DisenGCD(Ise+Rc)}& 0.7301 & 0.4235 & 0.7678  \\ %& 0.7509 & 0.4043 & 0.7839

        \textit{DisenGCD(Isc+Re)}  & 0.7333 & 0.4231 &
        0.7685  \\ %& 0.7506 & 0.4041 & 0.7834

        DisenGCD & \textbf{0.7335} & \textbf{0.4219} & \textbf{0.7723} \\ % & \textbf{0.7527} & \textbf{0.4041} & \textbf{0.7858}
        
        \bottomrule
\end{tabular}
\fi
\end{table}

\begin{figure}[t]
     \centering
      
           \subfloat[]{
           \includegraphics[width=0.46\linewidth]{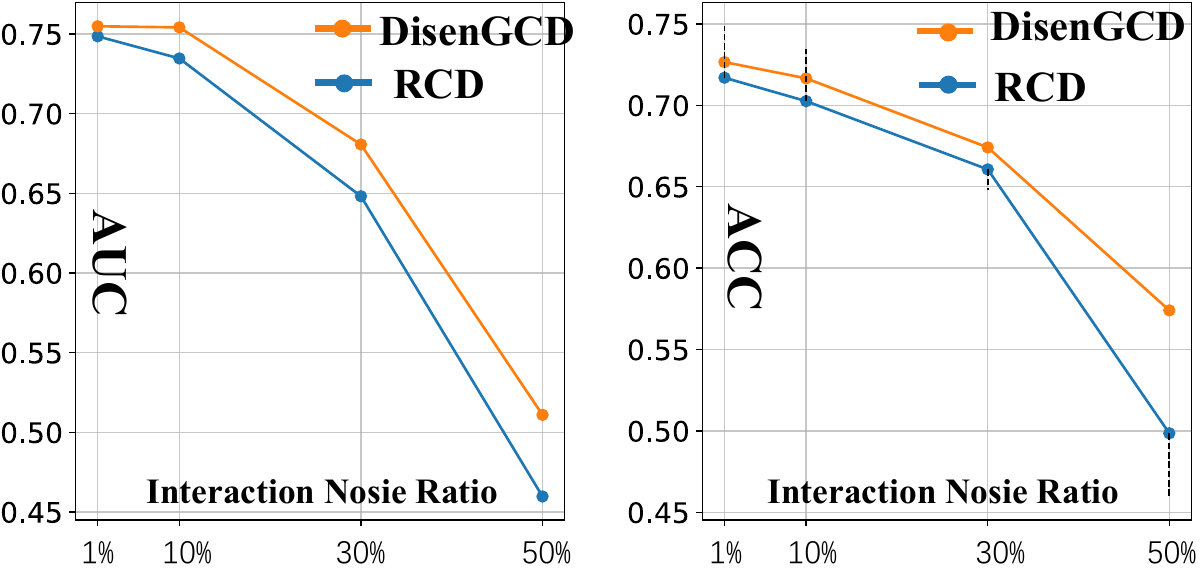}\label{figure:robust}
           }      
           \subfloat[]
           {\includegraphics[width=0.54\linewidth]{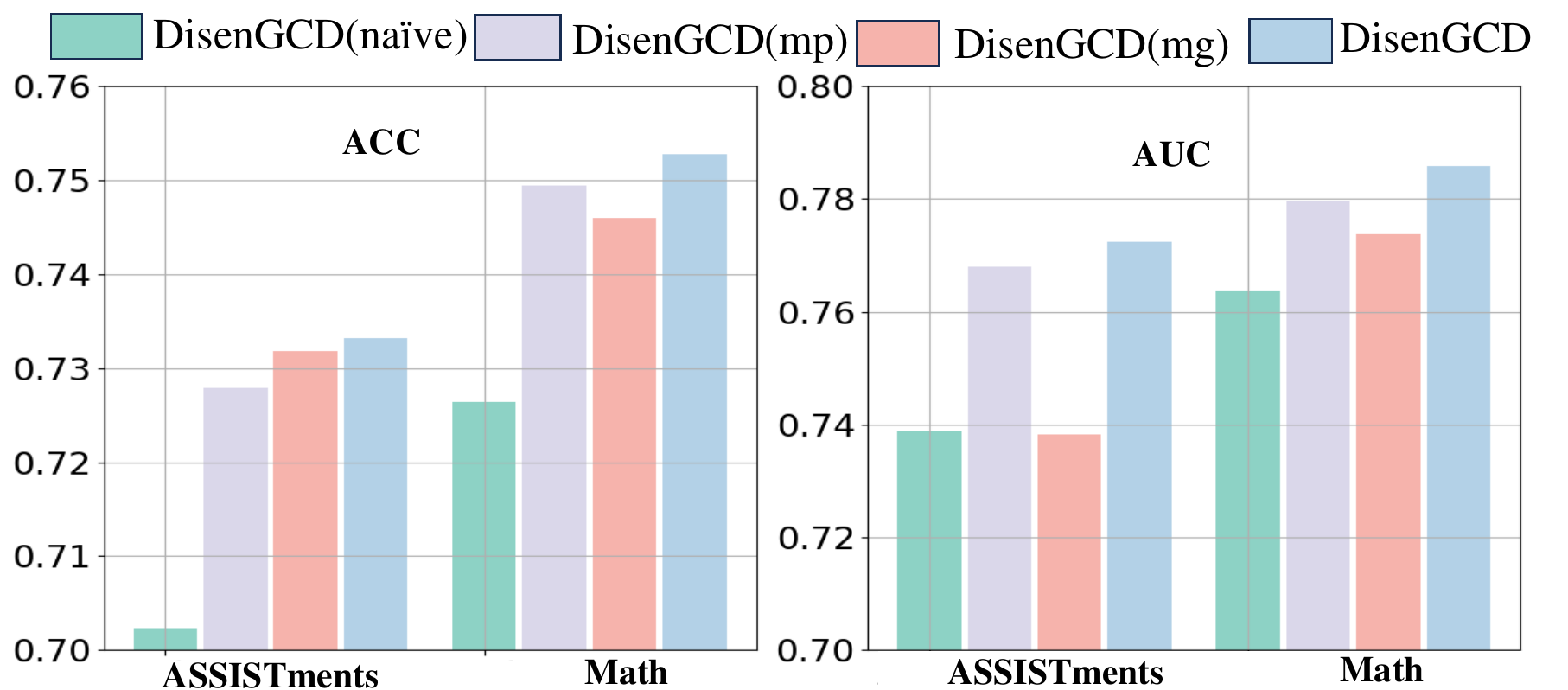}\label{performance compare}}      
      \caption{(a): Performance of RCD and DisenGCD under different noises. (b): Effectiveness of the meta multigraph learning module.}

\end{figure}

\subsection{Effectiveness of Disentangled Learning Framework of DisenGCD (\textbf{RQ2})}
\label{sec:RQ2}
To investigate  DisenGCD's robustness against interaction noise, 
we conducted robust experiments on the ASSISTments dataset under the ratio of 60\%/10\%/30\%, where a certain amount of noise interactions were added to each student in the training and validation datasets. 
Figure~\ref{figure:robust} presents the ACC and AUC of DisenGCD and RCD  under noise data of different percentages. 
 As the noise data increases, the performance leading of DisenGCD over RCD becomes more significant, which reaches maximal when noise data of 50\% was added.
That demonstrates  DisenGCD is more robust to student noise interactions than RCD, attributed to the disentangled learning framework of DisenGCD.
Similar observations can be drawn from \textbf{more experiments on other two datasets  in Appendix~\ref{sec:exp_robust}}.

To further analyze the framework effectiveness, four variants of DisenGCD were created:
 \textit{DisenGCD(I)} refers to learning three representations only on  the interaction graph $\mathcal{G_I}$; \textit{DisenGCD(Is+Rec)} refers to learning student representation on  $\mathcal{G_I}$, but learning exercise and concept ones on the relation graph $\mathcal{G_R}$;
\textit{DisenGCD(Ise+Rc)}  refers to learning student and exercise representations on $\mathcal{G_I}$ but learning 
concept one on $\mathcal{G_R}$;
while \textit{DisenGCD(Isc+Re)} learns  student and concept representations on $\mathcal{G_I}$ but learns 
exercise one on $\mathcal{G_R}$. 
Table~\ref{ablation experiment} compares the performance of RCD, DisenGCD, and its four variants on ASSISTments.
As can be seen,  the comparison between    \textit{DisenGCD(I)} and other variants 
indicates learning three representations in two disentangled graphs is more effective than in one unified graph, especially with  textit{DisenGCD(Is+Rec)};
the comparison between    \textit{DisenGCD(Is+Rec)} and DisenGCD indicates learning three representations in three disentangled graphs is more effective, further validating the above conclusion.
Finally, we can conclude that the proposed disentangled learning framework is effective in enhancing DisenGCD's performance and robustness.

\iffalse
\begin{figure}[t]
     \centering

      \caption{Effectiveness of the meta multigraph learning module.}

\end{figure}

\fi

\subsection{Effectiveness of Meta Multigraph Learning Module in DisenGCD (\textbf{RQ3})}

%如图3所示，三种模型在ASSIST和MATH两个数据集（70%的训练集，30%的测试集）中的比较结果。我们可以发现我们的MMCD效果在两个数据集中都取得了最好的效果,这说明模型中元多图聚合模块的有效性，它可以更好的对学生表征进行建模。除此之外，我们还可以发现单路径的元图聚合在math数据集中的效果最差，它说明我们仅仅考虑元图中权重最大的边可能不能够很好的聚合图中信息，且存在不稳定的问题
In Table~\ref{ablation experiment}, both  \textit{DisenGCD(I)} and RCD learn three types of representations in one unified graph, but  \textit{DisenGCD(I)} utilizes the meta multigraph aggregator. Therefore,  the comparison between  \textit{DisenGCD(I)} and RCD proves 
the devised meta multigraph learning module is effective in improving the model performance to some extent.

To further validate this, we created three variants of DisenGCD: 
\textit{DisenGCD(naive)} refers to the meta multigraph module replaced by the naive embedding (i.e., $\overline{\mathbf{S_i}}$ equal to $\mathbf{s}_i$ in Eq.(\ref{eq:embedding}));
\textit{DisenGCD(mp)} refers to predefining the meta multigraph module's propagation paths according to HAN~\cite{HAN};
\textit{DisenGCD(mg)} refers to this module's meta multigraph replaced by meta graph.
Figure~\ref{performance compare} presents the AUC and ACC values of DisenGDCN and its variants on two datasets under the ratio of 60\%/10\%/30\%.
As can be seen, the devised meta multigraph learning module enables DisenGDCN to hold significantly better performance than the naive embedding-based variant. 
The comparison results between  DisenGCD and \textit{DisenGCD(mg)} as well as \textit{DisenGCD(mp)} 
validates the effectiveness of using meta multigraph and the automatically learned propagation paths (i.e., learned meta multigraph). 
Thus, the effectiveness of the devised meta multigraph  module can be validated.

%可视化实验
\begin{figure}[t]

     \centering
     \subfloat[]{
           \includegraphics[width=0.55\linewidth]{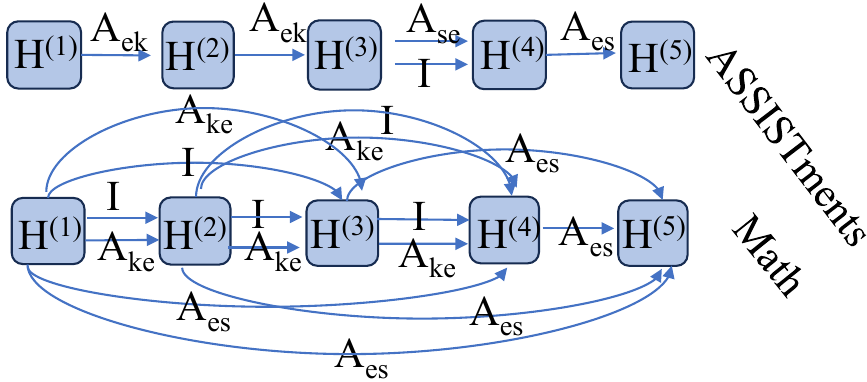}
      \label{fig:visual}
     }
     \subfloat[]{
           \includegraphics[width=0.45\linewidth]{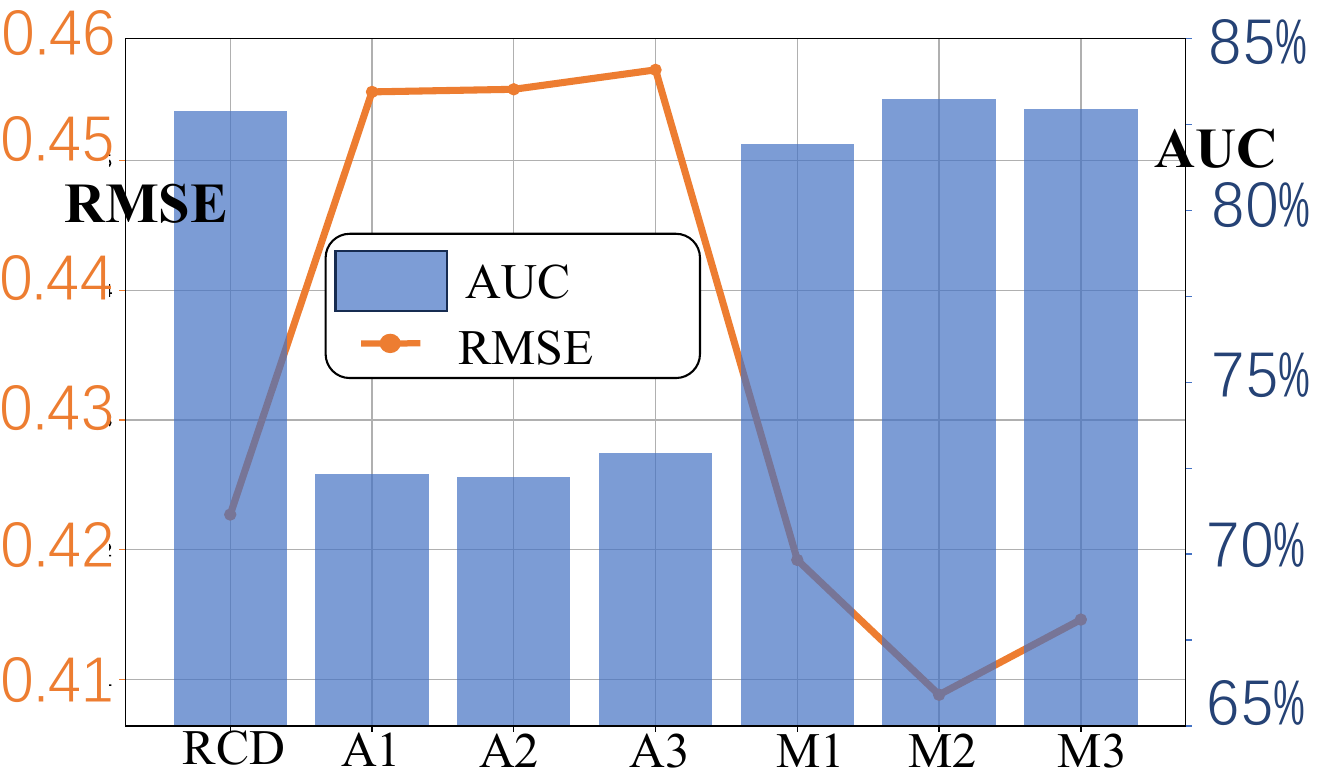}
      \label{fig:gener}
     }
\caption{(a): Visualization of learned meta multigraph on two datasets. (b): Generalization validation of six learned meta multigraphs.}

\end{figure}

\subsection{Visualization and Effectiveness of Learned Meta Multigraph~(\textbf{RQ4})}
The comparison between  \textit{DisenGCD(mp)} and DisenGCD has revealed the effectiveness of the learned meta multigraph on the target dataset.
Therefore,  an intuitive doubt naturally emerged:
Is the learned meta multigraph still effective on other datasets?
Before solving this, Figure~\ref{fig:visual} gives the structure visualization of two learned meta multigraphs. 
We can see two learned meta multigraphs hold two distinct structures. 
That may be because two datasets are a bit different regarding the relations between exercises and concepts, where exercises in Math contain only one concept while exercises in ASSISTments may contain more than one concept. Thus, another doubt naturally emerged: Is the learned meta multigraph on the target dataset ineffective on a different type of dataset?

To solve the doubts, we applied six learned meta multigraphs (A1, A2, A3, M1, M2, and M3)  to the SLP. A1-A3 and M1-M3 were learned by DisenGCD three times on ASSISTments and Math. 
 SLP  dataset is similar to Math, whose exercises only contain one concept.
Figure~\ref{fig:gener} summarizes the results of RCD and six DisenGCD variants that utilize six given meta multigraphs. As can be seen, DisenGCD's performance under M1-M3 is promising and better than RCD, while DisenGCD under A1-A3 performed poorly. 
The observation can answer the above doubts to some extent:  meta multigraphs learned by DisenGCD may be effective when the datasets to be applied are similar to target datasets.

In addition to the above four experiments, \textbf{more experiments} were executed to  validate  the devised diagnostic functions,  analyze the DisenGCD's parameter sensitivity, analyze its execution efficiency, 
and investigate the effectiveness of the employed GAT modules in \textbf{Appendix~\ref{sec:exp_diag}},
\textbf{Appendix~\ref{sec:exp_sensi}}, \textbf{Appendix~\ref{sec:efficiency}}, and \textbf{Appendix~\ref{sec:ablationGAT}}.

\section{Conclusion}
This paper proposed a meta multigraph-assisted disentangled graph learning framework for CD, called DisenGCD.
The proposed DisenGCD learned student, exercise, and concept representations on three disentangled graphs, respectively. It devised a meta multigraph module to learn student representation and employed two common GAT modules to learn exercise and concept representations.
Compared to SOTA CDMs on three datasets, the proposed DisenGCD exhibited highly better performance and showed high robustness against interaction noise. 

\newpage
	\section*{Acknowledgements}
	
This work was supported in part by the National Key R\&D Program of China (No.2018AAA0100100),  
 in part by the National Natural Science Foundation of China (No.62302010, No.62303013, No.62107001, No.62006053, No.61876162, No.62136008, No.62276001, No.U21A20512),  
 in part by China Postdoctoral Science Foundation (No.2023M740015), 
  in part by the Postdoctoral Fellowship Program (Grade B) of China Postdoctoral Science Foundation (No.GZB20240002),
  and in part by the  Anhui Province Key Laboratory of Intelligent Computing and Applications (No. AFZNJS2024KF01).

\bibliographystyle{plain}
\bibliography{nips24}

\begin{thebibliography}{10}

\bibitem{onlinecourses}
Ashton Anderson, Daniel Huttenlocher, Jon Kleinberg, and Jure Leskovec.
\newblock Engaging with massive online courses.
\newblock In {\em Proceedings of the 23rd international conference on World
  wide web}, pages 687--698, 2014.

\bibitem{beck2007difficulties}
Joseph Beck.
\newblock Difficulties in inferring student knowledge from observations (and
  why you should care).
\newblock In {\em Proceedings of the 13th International Conference of
  Artificial Intelligence in Education}, pages 21--30, 2007.

\bibitem{cheng2019dirt}
Song Cheng, Qi~Liu, Enhong Chen, Zai Huang, Zhenya Huang, Yiying Chen, Haiping
  Ma, and Guoping Hu.
\newblock Dirt: Deep learning enhanced item response theory for cognitive
  diagnosis.
\newblock In {\em Proceedings of the 28th ACM international conference on
  information and knowledge management}, pages 2397--2400, 2019.

\bibitem{dina}
Jimmy De~La~Torre.
\newblock Dina model and parameter estimation: A didactic.
\newblock {\em Journal of educational and behavioral statistics},
  34(1):115--130, 2009.

\bibitem{diffmg}
Yuhui Ding, Quanming Yao, Huan Zhao, and Tong Zhang.
\newblock Diffmg: Differentiable meta graph search for heterogeneous graph
  neural networks.
\newblock In {\em Proceedings of the 27th ACM SIGKDD Conference on Knowledge
  Discovery \& Data Mining}, pages 279--288, 2021.

\bibitem{feng2009addressing}
Mingyu Feng, Neil Heffernan, and Kenneth Koedinger.
\newblock Addressing the assessment challenge with an online system that tutors
  as it assesses.
\newblock {\em User Modeling and User-adapted Interaction}, 19(3):243--266,
  2009.

\bibitem{rcd}
Weibo Gao, Qi~Liu, Zhenya Huang, Yu~Yin, Haoyang Bi, Mu-Chun Wang, Jianhui Ma,
  Shijin Wang, and Yu~Su.
\newblock Rcd: Relation map driven cognitive diagnosis for intelligent
  education systems.
\newblock In {\em Proceedings of the 44th international ACM SIGIR conference on
  research and development in information retrieval}, pages 501--510, 2021.

\bibitem{gems}
Zhenyu Han, Fengli Xu, Jinghan Shi, Yu~Shang, Haorui Ma, Pan Hui, and Yong Li.
\newblock Genetic meta-structure search for recommendation on heterogeneous
  information network.
\newblock In {\em Proceedings of the 29th ACM International Conference on
  Information \& Knowledge Management}, pages 455--464, 2020.

\bibitem{lightgcn}
Xiangnan He, Kuan Deng, Xiang Wang, Yan Li, YongDong Zhang, and Meng Wang.
\newblock Lightgcn: Simplifying and powering graph convolution network for
  recommendation.
\newblock In {\em Proceedings of the 43rd International ACM SIGIR Conference on
  Research and Development in Information Retrieval}, SIGIR '20, page
  639–648, New York, NY, USA, 2020. Association for Computing Machinery.

\bibitem{PMMM}
Chao Li, Hao Xu, and Kun He.
\newblock Differentiable meta multigraph search with partial message
  propagation on heterogeneous information networks.
\newblock In {\em Proceedings of the AAAI Conference on Artificial
  Intelligence}, volume~37, pages 8518--8526, 2023.

\bibitem{liu2019exploiting}
Qi~Liu, Shiwei Tong, Chuanren Liu, Hongke Zhao, Enhong Chen, Haiping Ma, and
  Shijin Wang.
\newblock Exploiting cognitive structure for adaptive learning.
\newblock In {\em Proceedings of the 25th ACM SIGKDD International Conference
  on Knowledge Discovery \& Data Mining}, pages 627--635, 2019.

\bibitem{liu2024automated}
Sannyuya Liu, Qing Li, Xiaoxuan Shen, Jianwen Sun, and Zongkai Yang.
\newblock Automated discovery of symbolic laws governing skill acquisition from
  naturally occurring data.
\newblock {\em Nature Computational Science}, pages 1--12, 2024.

\bibitem{irt}
Frederique~M Lord.
\newblock {\em Applications of Item Response Theory to Practical Testing
  Problems}.
\newblock LAWRENCE ERLBAUM ASSCCIAATES, 1980.

\bibitem{lu2021slp}
Yu~Lu, Yang Pian, Ziding Shen, Penghe Chen, and Xiaoqing Li.
\newblock {SLP}: A multi-dimensional and consecutive dataset from {K-12}
  education.
\newblock In {\em Proceedings of the 29th International Conference on Computers
  in Education Conference}, volume~1, pages 261--266. Asia-Pacific Society for
  Computers in Education, 2021.

\bibitem{kscd}
Haiping Ma, Manwei Li, Le~Wu, Haifeng Zhang, Yunbo Cao, Xingyi Zhang, and
  Xuemin Zhao.
\newblock Knowledge-sensed cognitive diagnosis for intelligent education
  platforms.
\newblock In {\em Proceedings of the 31st ACM International Conference on
  Information \& Knowledge Management}, pages 1451--1460, 2022.

\bibitem{ma2024enhancing}
Haiping Ma, Changqian Wang, Hengshu Zhu, Shangshang Yang, Xiaoming Zhang, and
  Xingyi Zhang.
\newblock Enhancing cognitive diagnosis using un-interacted exercises: A
  collaboration-aware mixed sampling approach.
\newblock In {\em Proceedings of the AAAI Conference on Artificial
  Intelligence}, volume~38, pages 8877--8885, 2024.

\bibitem{ma2024hd}
Haiping Ma, Yong Yang, Chuan Qin, Xiaoshan Yu, Shangshang Yang, Xingyi Zhang,
  and Hengshu Zhu.
\newblock Hd-kt: Advancing robust knowledge tracing via anomalous learning
  interaction detection.
\newblock In {\em Proceedings of the ACM on Web Conference 2024}, pages
  4479--4488, 2024.

\bibitem{dlcat}
Haiping Ma, Yi~Zeng, Shangshang Yang, Chuan Qin, Xingyi Zhang, and Limiao
  Zhang.
\newblock A novel computerized adaptive testing framework with decoupled
  learning selector.
\newblock {\em Complex \& Intelligent Systems}, pages 1--12, 2023.

\bibitem{Ma2019DisentangledGC}
Jianxin Ma, Peng Cui, Kun Kuang, Xin Wang, and Wenwu Zhu.
\newblock Disentangled graph convolutional networks.
\newblock In {\em International Conference on Machine Learning}, 2019.

\bibitem{mirt}
Mark~D. Reckase.
\newblock {\em Multidimensional Item Response Theory}.
\newblock Springer New York, 2009.

\bibitem{gcdm}
Yu~Su, Zeyu Cheng, Jinze Wu, Yanmin Dong, Zhenya Huang, Le~Wu, Enhong Chen,
  Shijin Wang, and Fei Xie.
\newblock Graph-based cognitive diagnosis for intelligent tutoring systems.
\newblock {\em Knowledge-Based Systems}, 253:109547, 2022.

\bibitem{sun2023adversarial}
Jianwen Sun, Fenghua Yu, Sannyuya Liu, Yawei Luo, Ruxia Liang, and Xiaoxuan
  Shen.
\newblock Adversarial bootstrapped question representation learning for
  knowledge tracing.
\newblock In {\em Proceedings of the 31st ACM International Conference on
  Multimedia}, pages 8016--8025, 2023.

\bibitem{sun2024interpretable}
Jianwen Sun, Fenghua Yu, Qian Wan, Qing Li, Sannyuya Liu, and Xiaoxuan Shen.
\newblock Interpretable knowledge tracing with multiscale state representation.
\newblock In {\em Proceedings of the ACM on Web Conference 2024}, pages
  3265--3276, 2024.

\bibitem{GAT}
Petar Veli{\v{c}}kovi{\'c}, Guillem Cucurull, Arantxa Casanova, Adriana Romero,
  Pietro Lio, and Yoshua Bengio.
\newblock Graph attention networks.
\newblock {\em arXiv preprint arXiv:1710.10903}, 2017.

\bibitem{ncd}
Fei Wang, Qi~Liu, Enhong Chen, Zhenya Huang, Yuying Chen, Yu~Yin, Zai Huang,
  and Shijin Wang.
\newblock Neural cognitive diagnosis for intelligent education systems.
\newblock In {\em Proceedings of the AAAI conference on artificial
  intelligence}, volume~34, pages 6153--6161, 2020.

\bibitem{wang2022neuralcd}
Fei Wang, Qi~Liu, Enhong Chen, Zhenya Huang, Yu~Yin, Shijin Wang, and Yu~Su.
\newblock Neuralcd: a general framework for cognitive diagnosis.
\newblock {\em IEEE Transactions on Knowledge and Data Engineering},
  35(8):8312--8327, 2022.

\bibitem{wang2019mcne}
Hao Wang, Tong Xu, Qi~Liu, Defu Lian, Enhong Chen, Dongfang Du, Han Wu, and Wen
  Su.
\newblock Mcne: An end-to-end framework for learning multiple conditional
  network representations of social network.
\newblock In {\em Proceedings of the 25th ACM SIGKDD international conference
  on knowledge discovery \& data mining}, pages 1064--1072, 2019.

\bibitem{scd}
Shanshan Wang, Zhen Zeng, Xun Yang, and Xingyi Zhang.
\newblock Self-supervised graph learning for long-tailed cognitive diagnosis.
\newblock In {\em Proceedings of the AAAI Conference on Artificial
  Intelligence}, volume~37, pages 110--118, 2023.

\bibitem{Wang2020DisentangledGC}
Xiang Wang, Hongye Jin, An~Zhang, Xiangnan He, Tong Xu, and Tat-Seng Chua.
\newblock Disentangled graph collaborative filtering.
\newblock {\em Proceedings of the 43rd International ACM SIGIR Conference on
  Research and Development in Information Retrieval}, 2020.

\bibitem{HAN}
Xiao Wang, Houye Ji, Chuan Shi, Bai Wang, Yanfang Ye, Peng Cui, and Philip~S
  Yu.
\newblock Heterogeneous graph attention network.
\newblock In {\em The world wide web conference}, pages 2022--2032, 2019.

\bibitem{cdgk}
Xinping Wang, Caidie Huang, Jinfang Cai, and Liangyu Chen.
\newblock Using knowledge concept aggregation towards accurate cognitive
  diagnosis.
\newblock In {\em Proceedings of the 30th ACM International Conference on
  Information \& Knowledge Management}, pages 2010--2019, 2021.

\bibitem{Wang2020DisenHANDH}
Yifan Wang, Suyao Tang, Yuntong Lei, Weiping Song, Sheng Wang, and Ming Zhang.
\newblock Disenhan: Disentangled heterogeneous graph attention network for
  recommendation.
\newblock {\em Proceedings of the 29th ACM International Conference on
  Information \& Knowledge Management}, 2020.

\bibitem{DcRec}
Jiahao Wu, Wenqi Fan, Jingfan Chen, Shengcai Liu, Qing Li, and Ke~Tang.
\newblock Disentangled contrastive learning for social recommendation.
\newblock New York, NY, USA, 2022. Association for Computing Machinery.

\bibitem{courserecommendation}
Zhengyang Wu, Ming Li, Yong Tang, and Qingyu Liang.
\newblock Exercise recommendation based on knowledge concept prediction.
\newblock {\em Knowledge-Based Systems}, 210:106481, 2020.

\bibitem{DCCF}
Jiashu Zhao Dawei~Yin Xubin~Ren, Lianghao~Xia and Chao Huang.
\newblock Disentangled contrastive collaborative filtering.
\newblock {\em In Proceedings of the 46th International ACM SIGIR Conference on
  Research and Development in Information Retrieval}, 2023.

\bibitem{yang2024evolutionary}
Shangshang Yang, Haiping Ma, Ying Bi, Ye~Tian, Limiao Zhang, Yaochu Jin, and
  Xingyi Zhang.
\newblock An evolutionary multi-objective neural architecture search approach
  to advancing cognitive diagnosis in intelligent education.
\newblock {\em IEEE Transactions on Evolutionary Computation}, 2024.

\bibitem{yang2023designing}
Shangshang Yang, Haiping Ma, Cheng Zhen, Ye~Tian, Limiao Zhang, Yaochu Jin, and
  Xingyi Zhang.
\newblock Designing novel cognitive diagnosis models via evolutionary
  multi-objective neural architecture search.
\newblock {\em arXiv preprint arXiv:2307.04429}, 2023.

\bibitem{yang2024endowing}
Shangshang Yang, Linrui Qin, and Xiaoshan Yu.
\newblock Endowing interpretability for neural cognitive diagnosis by efficient
  kolmogorov-arnold networks.
\newblock {\em arXiv preprint arXiv:2405.14399}, 2024.

\bibitem{yang2024hybrid}
Shangshang Yang, Xiangkun Sun, Ke~Xu, Yuanchao Liu, Ye~Tian, and Xingyi Zhang.
\newblock Hybrid architecture-based evolutionary robust neural architecture
  search.
\newblock {\em IEEE Transactions on Emerging Topics in Computational
  Intelligence}, 2024.

\bibitem{yang2021gradient}
Shangshang Yang, Ye~Tian, Cheng He, Xingyi Zhang, Kay~Chen Tan, and Yaochu Jin.
\newblock A gradient-guided evolutionary approach to training deep neural
  networks.
\newblock {\em IEEE Transactions on Neural Networks and Learning Systems},
  33(9):4861--4875, 2021.

\bibitem{yang2022accelerating}
Shangshang Yang, Ye~Tian, Xiaoshu Xiang, Shichen Peng, and Xingyi Zhang.
\newblock Accelerating evolutionary neural architecture search via
  multifidelity evaluation.
\newblock {\em IEEE Transactions on Cognitive and Developmental Systems},
  14(4):1778--1792, 2022.

\bibitem{yang2023cognitive}
Shangshang Yang, Haoyu Wei, Haiping Ma, Ye~Tian, Xingyi Zhang, Yunbo Cao, and
  Yaochu Jin.
\newblock Cognitive diagnosis-based personalized exercise group assembly via a
  multi-objective evolutionary algorithm.
\newblock {\em IEEE Transactions on Emerging Topics in Computational
  Intelligence}, 2023.

\bibitem{ENAS-KT}
Shangshang Yang, Xiaoshan Yu, Ye~Tian, Xueming Yan, Haiping Ma, and Xingyi
  Zhang.
\newblock Evolutionary neural architecture search for transformer in knowledge
  tracing.
\newblock In {\em Thirty-seventh Conference on Neural Information Processing
  Systems}, 2023.

\bibitem{yang2023evolutionary}
Shangshang Yang, Cheng Zhen, Ye~Tian, Haiping Ma, Yuanchao Liu, Panpan Zhang,
  and Xingyi Zhang.
\newblock Evolutionary multi-objective neural architecture search for
  generalized cognitive diagnosis models.
\newblock In {\em 2023 5th International Conference on Data-driven Optimization
  of Complex Systems (DOCS)}, pages 1--10. IEEE, 2023.

\bibitem{yin2024dataset}
Mingjia Yin, Hao Wang, Wei Guo, Yong Liu, Suojuan Zhang, Sirui Zhao, Defu Lian,
  and Enhong Chen.
\newblock Dataset regeneration for sequential recommendation.
\newblock In {\em Proceedings of the 30th ACM SIGKDD Conference on Knowledge
  Discovery and Data Mining}, pages 3954--3965, 2024.

\bibitem{yin2024entropy}
Mingjia Yin, Chuhan Wu, Yufei Wang, Hao Wang, Wei Guo, Yasheng Wang, Yong Liu,
  Ruiming Tang, Defu Lian, and Enhong Chen.
\newblock Entropy law: The story behind data compression and llm performance.
\newblock {\em arXiv preprint arXiv:2407.06645}, 2024.

\bibitem{yu2024rigl}
Xiaoshan Yu, Chuan Qin, Dazhong Shen, Shangshang Yang, Haiping Ma, Hengshu Zhu,
  and Xingyi Zhang.
\newblock Rigl: A unified reciprocal approach for tracing the independent and
  group learning processes.
\newblock In {\em Proceedings of the 30th ACM SIGKDD Conference on Knowledge
  Discovery and Data Mining}, pages 4047--4058, 2024.

\bibitem{ecd}
Yuqiang Zhou, Qi~Liu, Jinze Wu, Fei Wang, Zhenya Huang, Wei Tong, Hui Xiong,
  Enhong Chen, and Jianhui Ma.
\newblock Modeling context-aware features for cognitive diagnosis in student
  learning.
\newblock In {\em Proceedings of the 27th ACM SIGKDD Conference on Knowledge
  Discovery \& Data Mining}, pages 2420--2428, 2021.

\end{thebibliography}

%%%%%%%%%%%%%%%%%%%%%%%%%%%%%%%%%%%%%%%%%%%%%%%%%%%%%

\newpage
\begin{appendices}
\section{More details about Proposed Method and Datasets}
\subsection{Notations Summary}\label{sec:notations}
For  convenient  reading and understanding, we have summarized all the notations discussed in the paper in the following two tables. 
Table~\ref{tab:decopling} summarizes all the notations in the three disentangled graphs, while Table~\ref{tab:meta multigraph} describes the notations of the meta multigraph.
\begin{table}[ht]
  \caption{Notation of disentangled graphs.}
    \centering
      \renewcommand{\arraystretch}{1.5}
    \begin{tabular}{c|c}
       \toprule
          Notation &  Description \\
          \hline
          $\mathcal{G_I}$ & the student-exercise-concept interaction  graph \\
          \hline
          $\mathbf{s}_{i}$ & initial embedding of student in $\mathcal{G_I}$ \\
          \hline
          $\mathbf{e}_{j}^I$ & initial embedding of exercise in $\mathcal{G_I}$ \\
          \hline
          $\mathbf{c}_{k}^I$ & initial embedding of concept in $\mathcal{G_I}$ \\
          \hline
         $\overline{\mathbf{S}_i}$ &  the final representation of the updated student node \\
         \hline
         $\mathcal{G_R}$ &  the exercise-concept relation Graph  \\
         \hline
         $\mathbf{e}_{j}^R$ &  initial embedding of exercise in $\mathcal{G_R}$ \\
         \hline
         $\mathbf{c}_{k}^R$ & initial embedding of concept in $\mathcal{G_R}$ \\
         \hline
         $\overline{\mathbf{E}_j}$ & the final representation of the updated exercise node \\
         \hline
         $\mathcal{G_D}$    & the concept dependency graph    \\
         \hline
         $\mathbf{c}_{k}^D$ & initial embedding of concept in $\mathcal{G_D}$    \\
         \hline
          $\overline{\mathbf{C}_k}$ & the final representation of the updated concept node  \\
          \hline
    \end{tabular}
    \label{tab:decopling}
\end{table}
\begin{table}[ht]
  \caption{Notation of meta multigraph.}
    \centering
  \renewcommand{\arraystretch}{1.5}
    \begin{tabular}{c|c}
       \toprule
          Notation &  Description \\
          \hline
          $\mathcal{MG}$ & meta multigraph \\
          \hline
          $\mathbf{H^{(p)}}$ & the p-th hyper node in meta multigraph \\
          \hline
          $\mathcal{AP}$ & edges between hyper nodes(propagation paths)  \\
           \hline
          $AP_{uv}$ & edges between $\mathbf{H^{(u)}}$ and   $\mathbf{H^{(v)}}$ \\
          \hline
          $HR$ & the set of propagation path types \\
          \hline
          $A_{se}$ & the propagation path from student to exercise \\
          \hline
          $A_{es}$ & the propagation path from exercise to student \\
          \hline
          $A_{ke}$ & the propagation path from concept to exercise \\
          \hline
          $A_{ek}$ & the propagation path from exercise to concept \\
          \hline
          $A_{k\hat{k}}$ & the propagation path from concept to concept \\
          \hline
          ${I}$  & information is not updated to propagate to the next hyper-node \\
          \hline
          $zero$  &  no information propagation between two hyper nodes \\
          \hline
          $hr_a^{wei}$  & the weight of each type propagation path in $AP_{uv}$  \\
          \hline
          $\tau^{(u, v)}$ & the threshold  of candidate propagation paths in$AP_{uv}$  \\
          \hline  
          \end{tabular}
    \label{tab:meta multigraph}
\end{table}

\subsection{Statistics of Datasets}\label{sec:data_details}
We evaluated our method on three real-world datasets: ASSISTments, Math and SLP, which both provide student-exercise interaction records and the exercise-knowledge concept relational matrix. 
\begin{itemize}
    \item $\mathbf{ASSISTments}$ is a public dataset collected by the assistant online tutoring systems in the 2009-2010 acadaemic year.
    \item $\mathbf{Math}$ is a private dataset collected by a well-known online learning platform that contains math practice records and test records for elementary and middle school students.
    \item $\mathbf{SLP}$ is another publicdata set that collects data on learners' performance in eight different subjects over three years of study, including maths, English, physics, chemistry, biology, history and geography.
\end{itemize}
 Notably, in both the Math dataset and the SLP dataset, they provide relationships between concepts. For these three datasets, we filtered students with fewer than 15 answer records to ensure there was enough data for the learning of the model.
We compared our model with five previous diagnostic models, including two classic models based on educational psychology, DINA~\cite{dina} and MIRT~\cite{mirt}, two neural network-based models, NCD~\cite{ncd} and ECD~\cite{ecd}, and a graph-based diagnostic model, RCD~\cite{rcd}. 
\begin{itemize}
    \item $\mathbf{DINA}$~\cite{dina} is a classical CDM, which uses a binary variable to characterize whether and whether a student has mastered a specific concept.
    \item $\mathbf{MIRT}$~\cite{mirt} is an extension of the irt model, which uses multidimensional vectors to characterize students' abilities and the difficulty of exercises, and uses linear functions to model the interactions.
    \item  $\mathbf{NCD}$~\cite{ncd} is one of the recent CDMs based on deep learning, which uses neural networks to model higher-order student-exercise complex interactions.
    \item $\mathbf{ECD}$~\cite{ecd} incorporates the impact of the educational environment in students' historical answer records into the diagnostic model to achieve diagnostic enhancement. Here due to the lack of educational background, we use random initialization vectors to characterize cognitive states.
    \item $\mathbf{RCD}$~\cite{rcd} is one of the most advanced models, which introduces the relationship between concepts into cognitive diagnosis and models the relationship through a graph structure.
\end{itemize}

\section{Additional Experiments}

 %Table generated by Excel2LaTeX from sheet 'Sheet1'
	 \begin{table*}[t]  
	 	\centering 
	\caption{ Performance comparison of  recent CDMs (SCD, KaNCD, KSCD) and  DisenGCD on the Math dataset.}  
				\renewcommand{\arraystretch}{1.1}

	 	 \setlength{\tabcolsep}{1.1mm}{
	 	\begin{tabular}{c|rrrr}    \toprule    \multirow{2}[2]{*}{\textbf{Method/Metric}} & \multicolumn{1}{c}{\multirow{2}[2]{*}{\textbf{SCD}}} & \multicolumn{1}{c}{\multirow{2}[2]{*}{\textbf{KaNCD}}} & \multicolumn{1}{c}{\multirow{2}[2]{*}{\textbf{KSCD}}} & \multicolumn{1}{c}{\multirow{2}[2]{*}{\textbf{DisenGCD}}}\\
   &       &       &       &        \\   
   \midrule    \textbf{ACC} & 0.7546 & 0.7519 & 0.7549 & \textbf{0.7582} \\    
   \textbf{RMSE} & 0.4025 & 0.4050 & 0.4040 & \textbf{0.4004}  \\  
   \textbf{AUC} & 0.7882 & 0.7854 & 0.7890 & \textbf{0.7932} \\    
   \bottomrule    \end{tabular}% 
	 }

	 	 \label{tab:sotaCDM1}%
	 	 \end{table*}%

\subsection{Experiments on  Dataset with Different Levels of Sparsity}\label{sec:exp_sparse}

\begin{table}[htbp]
  \centering
    \setlength{\tabcolsep}{.8mm}{
  \caption{Performance comparison between DisenGCD, NCD, and RCD under sparse interaction data.}
    \begin{tabular}{c|l|ccc|ccc|ccc}
    \toprule
    \multirow{2}[4]{*}{Datasets} & \multicolumn{1}{c|}{\multirow{2}[4]{*}{Models}} & \multicolumn{3}{c|}{5\% sparsity} & \multicolumn{3}{c|}{10\% sparsity} & \multicolumn{3}{c}{20\% sparsity} \\
\cmidrule{3-11}          &       & ACC   & RMSE  & AUC   & ACC   & RMSE  & AUC   & ACC   & RMSE  & AUC \\
    \midrule
    \multirow{3}[2]{*}{ASSISTments} & NCD   & 0.7216 & 0.4337 & 0.7487 & 0.7170  & 0.4365 & 0.7455 & 0.7146 & 0.4471 & 0.7398 \\
          & RCD   & 0.7264 & 0.4273 & 0.7636 & 0.7245 & 0.4295 & 0.7610  & 0.7241 & 0.4289 & 0.7569 \\
          & DisenGCD & 0.7282 & 0.4248 & 0.7676 & 0.7278 & 0.4254 & 0.7660  & 0.7273 & 0.4256 & 0.7631 \\
    \midrule
    \multirow{3}[2]{*}{Math} & NCD   & 0.7399 & 0.4122 & 0.7674 & 0.7383 & 0.4148 & 0.7651 & 0.7352 & 0.4162 & 0.7629 \\
          & RCD   & 0.7496 & 0.4071 & 0.7790  & 0.7481 & 0.4104 & 0.7781 & 0.7476 & 0.4075 & 0.7771 \\
          & DisenGCD & 0.7526 & 0.4030  & 0.7891 & 0.7503 & 0.4049 & 0.7832 & 0.7485 & 0.4058 & 0.7812 \\
    \bottomrule
    \end{tabular}%
    }
  \label{tab:spar}%
\end{table}%

As shown in Section \ref{sec:RQ1}, the proposed DisenGCD shows better performance  than RCD even when the splitting ratios change, which indicates the better robustness of DisenGCD to 
different-sparsity data. 

To further verify whether the proposed DisenGCD is robust against sparse  data (i.e., missing data or sparse interaction patterns),  we conducted corresponding experiments on ASSISTments and Math datasets under the splitting setting of 60\%/ 10\% / 30\%.
In the experiments, for each dataset,
we randomly deleted 5\%, 10\%, and 20\% of the students' answer records (i.e., the interaction data) in the training sets, respectively, thus providing three variant training sets;
the proposed DisenGCD was compared with NCD and RCD, 
and their overall performance comparisons were summarized in Table~\ref{tab:spar}.

As can be seen, under which deleting ratio, the proposed DisenGCD exhibits the best performance, and its performance does not drop very significantly  when the deleting ratio increases. 
Even when deleting 20\% of students' interaction data, DisenGCD still achieves an AUC value of 0.7812, which is greatly higher than NCD and RCD.
It demonstrates that the proposed DisenGCD is  effective on sparse interaction data, 
which is mainly attributed to the devised meta-multigraph learning module. 
That module enables the DisenGCD to access and use lower-order exercise latent representations, thus providing more accurate and robust students' representations, especially when partial data is lacking.

In summary, the proposed DisenGCD is effective and more robust against sparse interaction data, which benefits from the devised meta-multigraph learning module, 
and thus the effectiveness  of the  meta-multigtaph learning module can be indirectly demonstrated.

\subsection{More Experiments for   Robustness Validation}\label{sec:exp_robust}

\begin{figure}[t]
     \centering
      \includegraphics[width=1.\linewidth]{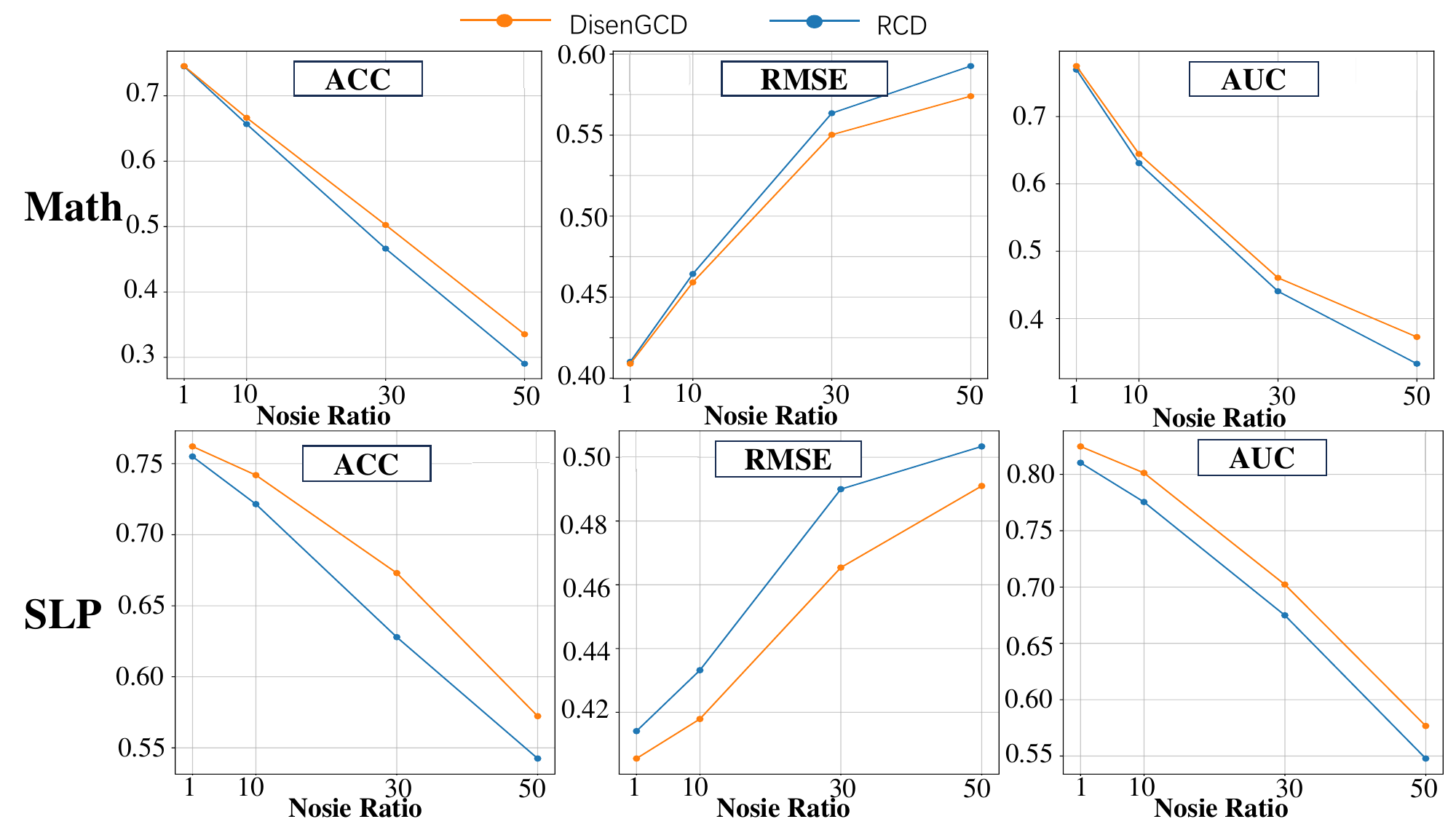}
      \caption{Futher Robustness Validation of DisenGCD. }
      \label{fig:robust}
\end{figure}

In Section~\ref{sec:RQ2}, we have validated the robustness of the proposed DisenGCD against interaction noise on the ASSISTments dataset.
To further show its robustness superiority,  
we executed the same experiments on other two datasets, i.e., Math and SLP, under the same settings,  where a certain ratio of noise
interactions were added to each student in the training and validation datasets.
 Figure~\ref{fig:robust} presents the results of DisenGCD and RCD obtained under different proportions of noise: the orange polyline denotes DsienGCD's results and the blue polyline denotes  RCD's results.

As can be seen, the proposed  DisenGCD always performs better than RCD in both datasets under different noise ratios. 
Besides,  as the noise ratio increases, the performance leading of the proposed DisenGCD to RCD becomes more significant.  
That further demonstrates the proposed  DisenGCD is more robust against student interaction noise and thus validates 
the effectiveness of the proposed disentangled graph learning framework.

\subsection{Validation of the Devised   Diagnostic Function}\label{sec:exp_diag}
\begin{table}[t]
  \centering
  \caption{Validation of the devised   diagnostic function on Math dataset.}
  \small
  \setlength{\tabcolsep}{.45mm}{
    \begin{tabular}{|l|ccc|ccc|ccc|ccc|}
    \toprule
    Splitting Ratios & \multicolumn{3}{c|}{70\%/10\%/20\%} & \multicolumn{3}{c|}{60\%/10\%/30\%} & \multicolumn{3}{c|}{50\%/10\%/40\%} & \multicolumn{3}{c|}{40\%/10\%/50\%} \\
    \cline{2-13}
         Diagnostic Functions  & \multicolumn{1}{l}{ACC} & \multicolumn{1}{l}{RMSE} & \multicolumn{1}{l|}{AUC} & \multicolumn{1}{l}{ACC} & \multicolumn{1}{l}{RMSE} & \multicolumn{1}{l|}{AUC} & \multicolumn{1}{l}{ACC} & \multicolumn{1}{l}{RMSE} & \multicolumn{1}{l|}{AUC} & \multicolumn{1}{l}{ACC} & \multicolumn{1}{l}{RMSE} & \multicolumn{1}{l|}{AUC} \\
    \midrule
    MIRT  & 0.7340 & 0.4256 & 0.7542 & 0.7279 & 0.4291 & 0.7479 & 0.7227 & 0.4299 & 0.7497 & 0.7229 & 0.4335 & 0.7427 \\ 
    NCD  & 0.7447 & 0.4084 & 0.7756 & 0.7418 & 0.4109 & 0.7706 & 0.7424 & 0.4119 & 0.7660 & 0.7394 & 0.4157 & 0.7604 \\
    RCD & 0.7534 & 0.4034 & 0.7866 & 0.7501 & 0.4078 & 0.7806 & 0.7489 & 0.4074 & 0.7751 & 0.7446 & 0.4100 & 0.7724 \\
    \hline
    DisenGCD-MIRT & 0.7396   & 0.4095  & 0.7713  & 0.7385  & 0.4103  & 0.7705  & 0.7378  & 0.4118  & 0.7689  & 0.7367  & 0.4138  & 0.7656  \\
    DisenGCD-NCD & 0.7501  & 0.4057  & 0.7796  & 0.7464  & 0.4076  & 0.7763  & 0.7430  & 0.4099  & 0.7714  & 0.7424  & 0.4127  & 0.7658   \\
    DisenGCD-RCD & 0.7548  & 0.4045  & 0.7874  & 0.7507  & 0.4066  & 0.7828  & 0.7482  & 0.4073  & 0.7795  & 0.7470  & 0.4098  & 0.7767\\
    \hline
    \textbf{DisenGCD} & \textbf{0.7582}  & \textbf{0.4004}  & \textbf{0.7932}  & \textbf{0.7527}  & \textbf{0.4039}  & \textbf{0.7867}  & \textbf{0.7513}  & \textbf{0.4052}  & \textbf{0.7832}  & \textbf{0.7479}  & \textbf{0.4076}  & \textbf{0.7802}  \\
    \bottomrule
    \end{tabular}%
}
  \label{tab:otherformulas}%
\end{table}%

In this paper, in addition to the disentangled learning framework and the devised meta multigraph module, we also designed a novel diagnostic function to adopt them. 
To validate its effectiveness, we compared it with the diagnostic functions of MIRT, NCD, and RCD  on the Math dataset, where four splitting settings were considered. 
Table~\ref{tab:otherformulas} summarizes their overall performance regarding AUC, ACC, and RMSE.

We can get the following two observations.
Firstly,  the comparisons of MIRT and DisenGCD-MIRT, NCD and DisenGCD-NCD, as well as
RCD and DisenGCD-RCD show that the representations learned by the proposed DisenGCD are effective.
Secondly,  the performance superiority of DisenGCD to other variants 
indicates the devised diagnostic function is effective. 
To sum up, the effectiveness of the  devised diagnostic function is validated.

\subsection{Hyperarameter Sensitivity Analysis}\label{sec:exp_sensi}
\begin{figure}[h]
     \centering
      \includegraphics[width=1.0\linewidth]{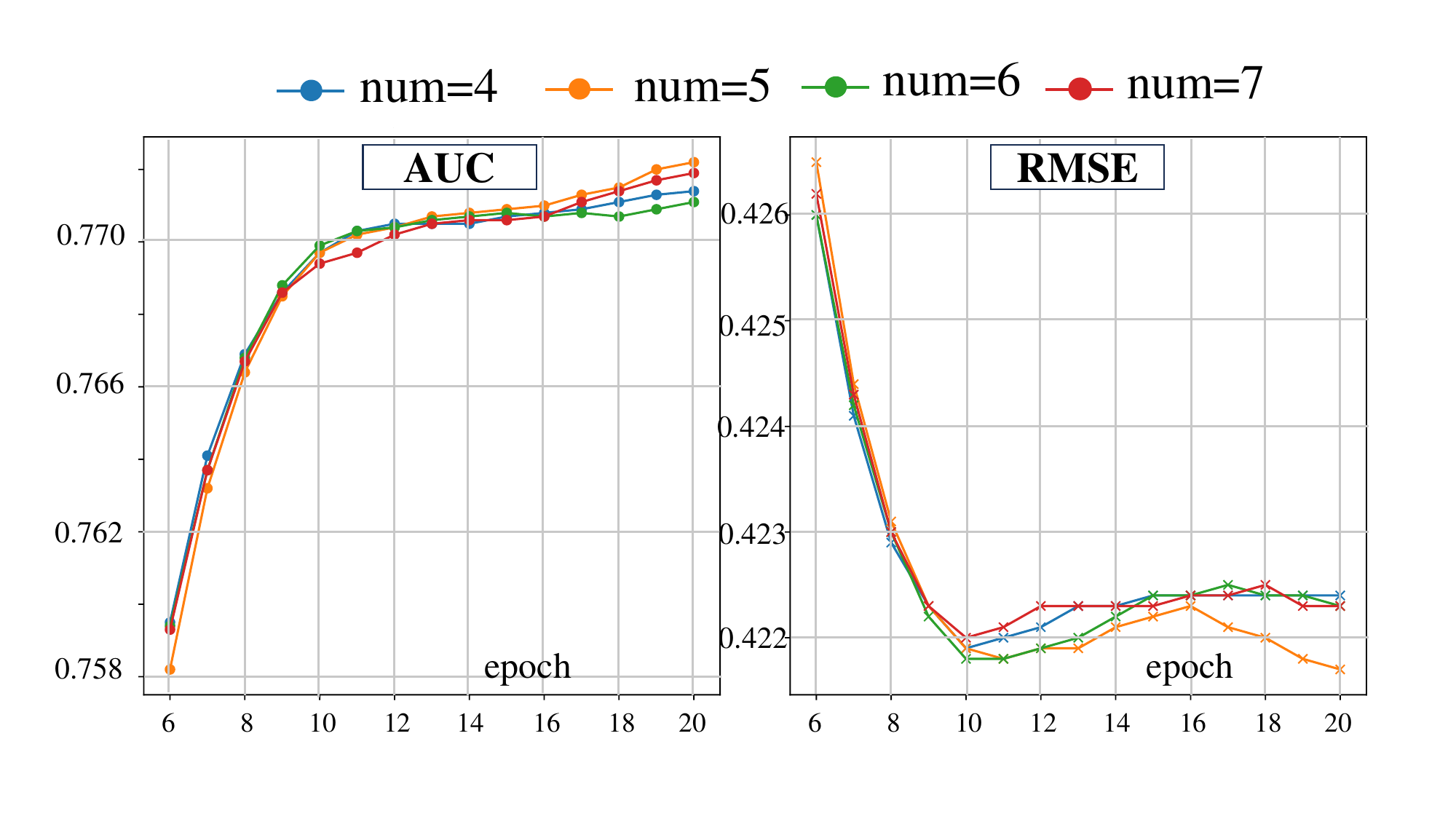}
      \caption{The impact of the number of hyper nodes in the meta-multigraph on the ASSISTments dataset }
      \label{fig:param}
\end{figure}
\begin{figure}[h]
  \centering
    \includegraphics[width=1.\textwidth]{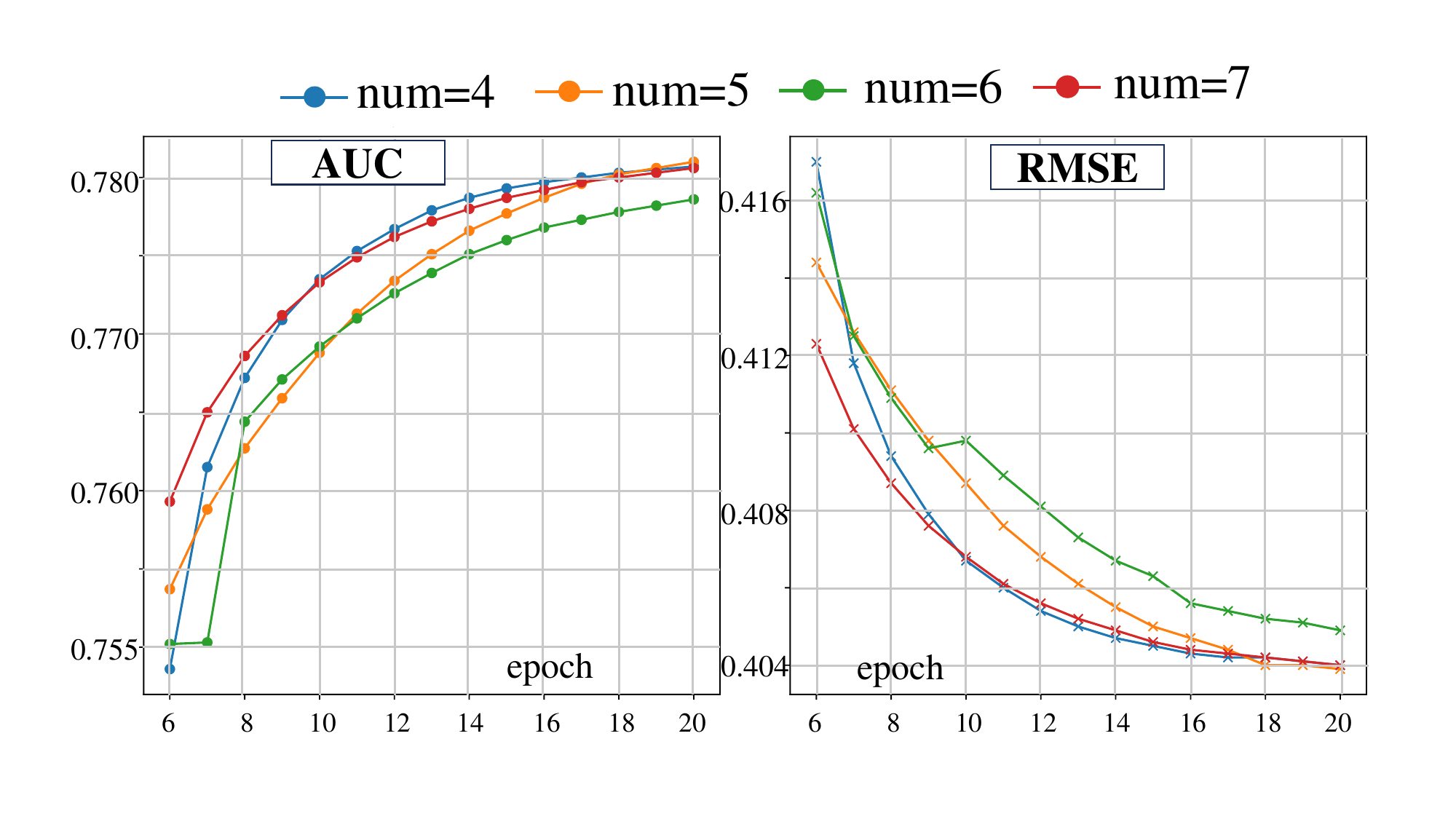}
    \caption{The impact of the number of hyper nodes in the meta-multigraph on the Math dataset}
    \label{fig:graph3}
\end{figure}

\begin{figure}[t]
\centering
     \subfloat{
           \includegraphics[width=0.45\linewidth,height=0.45\linewidth]{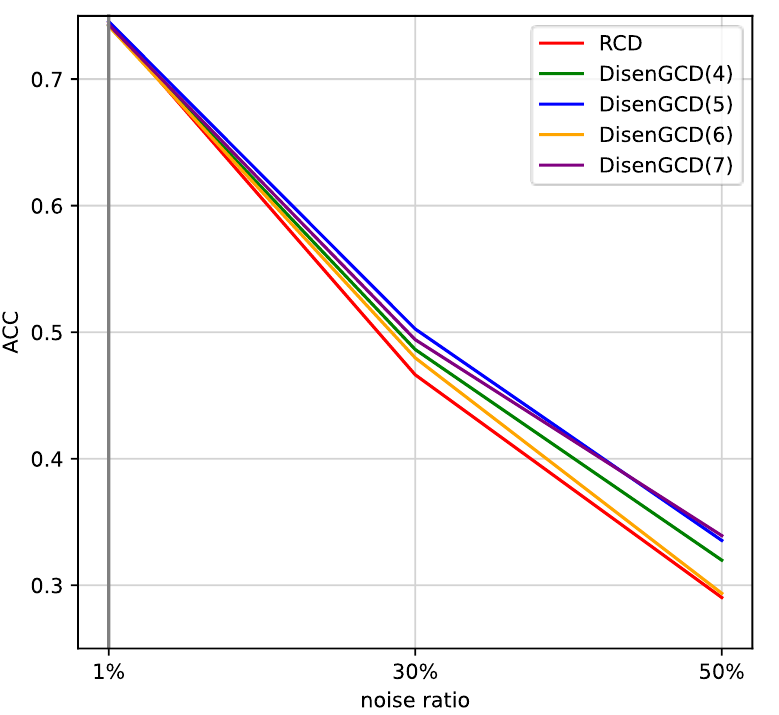}

     }
     \subfloat{
           \includegraphics[width=0.45\linewidth,height=0.45\linewidth]{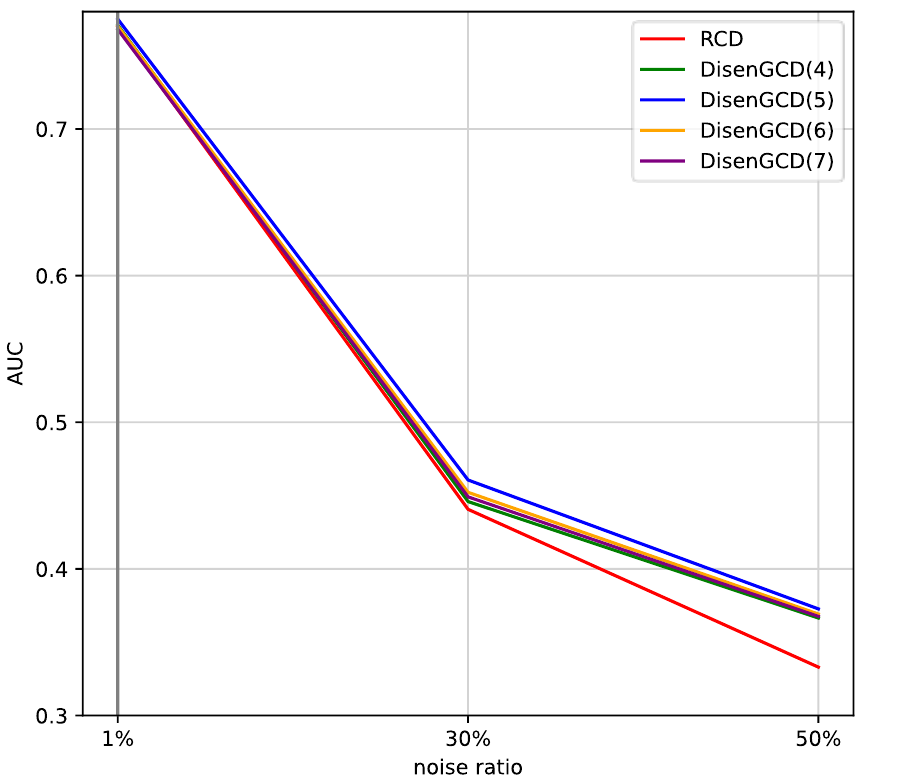}

     }
\caption{Robustness of DisenGCD under different $P$ to different ratios of interaction noise (on the Math dataset). }      \label{fig:gener1}
\end{figure}

For the proposed DisenGCD, there is an important hyperparameter, i.e., the number of hyper-nodes in the meta multigraph. 
To investigate its influence on theDisenGCD, 
the hyper-node number was set to 4, 5, 6, and  7, respectively, and we executed the experiments on the ASSISTments and SLP datasets. 
Figures~\ref{fig:param} and~\ref{fig:graph3} plot the results regarding AUC and RMSE.

 As can be seen in both two datasets, 
 the DisenGCD can achieve optimal results in terms of  RMSE and AUC, when the number of hyper-nodes is set to 5.
When the number of hyper-nodes is less than 5, the meta multigraph contains too few propagation paths, adversely affecting the learning of student representations. 
This results in suboptimal performance compared to the scenario where the number of hyper-nodes is 5. 
Conversely, when the number of hyper-nodes is greater than 5, 
there is an abundance of propagation paths, making the aggregated information overly complex and hindering effective student representation learning. 
Besides, more hyper-nodes cause more computational complexity.
Therefore, this paper set the number of hyper-nodes to 5 for the proposed disenGCD.

Furthermore, we aim to explore the influence of the number of hyper-nodes on the robustness of the proposed DisenGCD. 
Therefore, we also executed the experiments on Math datasets under different ratios of noise interaction(1\%,30\%,50\%). Figures~\ref{fig:gener1} plot the results regarding AUC and ACC. 
As can be seen, when the number of hyper nodes is equal to 5, the robustness of DisenGCD is the most promising, whose performance under different ratios of noises is balanced better than RCD.

% Table generated by Excel2LaTeX from sheet 'Sheet1'
\begin{table}[t]
	\centering
	
				\renewcommand{\arraystretch}{1.2}
			
		\begin{tabular}{c|lc}

		\toprule
		Datasets & Models & \multicolumn{1}{l}{Inference time (seconds)} \\
		\midrule
		\multirow{2}[2]{*}{ASSISTments} & RCD   & 0.07829 \\
		& DisenGCD & 0.01661 \\
		\midrule
		\multirow{2}[2]{*}{Math} & RCD   & 0.10888 \\
		& DisenGCD & 0.00791 \\
		\bottomrule
	\end{tabular}%

	\begin{tabular}{c|ccc|ccc}
		\toprule
		\multirow{2}[4]{*}{Models} & \multicolumn{3}{c|}{ASSISTments} & \multicolumn{3}{c}{Math} \\
		\cmidrule{2-7}          & RMSE  & AUC   & Training time(s) & RMSE  & AUC   & Training time(s) \\
		\midrule
		RCD   & 0.4245 & 0.7687 & 27161 & 0.4078 & 0.7806 & 22164 \\
		DisenGCD(4) & 0.4224 & 0.7714 & 17880 & 0.404 & 0.7857 & 3045 \\
		DisenGCD(5) & 0.4217 & 0.7722 & 18135 & 0.4039 & 0.786 & 3346 \\
		DisenGCD(6) & 0.4223 & 0.7711 & 18924 & 0.4049 & 0.7836 & 3675 \\
		DisenGCD(6) & 0.4223 & 0.7719 & 20004 & 0.404 & 0.7856 & 4681 \\
		\bottomrule
	\end{tabular}%
        \caption{\textbf{Upper}: Inference time comparison between RCD and DisenGCD. \textbf{Lower}:  Performance and computational efficiency (regarding   training runtime seconds) of DisenGCD under different $P$ (equal to 4, 5, ,6, and 7). }
	\label{tab:11}%
\end{table}%
\begin{table}[htbp]
  \centering
    \begin{tabular}{c|cccc}
    \toprule
    \multicolumn{1}{l|}{Method/Metric} & DisenGCD(GCN) & DisenGCD(GraphSage) & DisenGCD(HAN) & DisenGCD \\
    \midrule
    ACC   & 0.7428 & 0.7543 & 0.7435 & \textbf{0.7582} \\
    RMSE  & 0.4112 & 0.402 & 0.411 & \textbf{0.4004} \\
    AUC   & 0.7639 & 0.7881 & 0.7647 & \textbf{0.7932} \\
    \bottomrule
    \end{tabular}%
    \caption{Performance comparison of variants of DisenGCD using different GNNs, and  DisenGCD on the Math dataset}
  \label{tab:sotaCDM}%
\end{table}%

\subsection{Efficiency Analysis of DisenGCD}\label{sec:efficiency}

To investigate the computational efficiency of the proposed approach, we have compared it with RCD on ASSISTments and Math datasets in terms of model inference time and training time. Table~\ref{tab:11} presents the overall comparison, where the time of the proposed DisenGCD under different hyperparameters P is also reported.

As can be seen, DisenGCD's inference time is better than RCD's. This indicates that the proposed DisenGCD is more efficient than RCD, further signifying that it is promising to extend DisenGCD to dynamic CD. It can be seen from Table III:Lower: although a larger P will make DisenGCD take more time to train the model, the proposed DisenGCD achieves the best performance on both two datasets when P=5 and its runtime does not increase too much, which is much better than RCD.
% Table generated by Excel2LaTeX from sheet 'Sheet1'

\subsection{Ablation Experiments on Graph Representation Learning}\label{sec:ablationGAT}

To verify the superiority of the employed GAT  on the proposed DisenGCD, three GNNs (GCN, GraphSage, and HAN) are used to replace the GAT in the proposed DisenGCD, which are termed DisenGCD(GCN), DisenGCD(GraphSage), and DisenGCD(HAN), respectively. Then, we compared the three approaches with the proposed DisenGCD on the Math dataset. The comparison of them is presented in Table~\ref{tab:sotaCDM}
As shown in Table~\ref{tab:sotaCDM}, the GCN-based DisenGCD achieves the worst performance, followed by DisenGCD(HAN). The GraphSage-based DisenGCD holds a competitive performance to yet is still worse than the proposed DisenGCD (based on GAT).

The above results show the GAT is a suitable and optimal choice among these four GNNs for the proposed DisenGCD, but we think the GAT may not be the best choice because there exist other types of GNNs that can be integrated with DisenGCD, showing better performance.

In summary, it is reasonable and effective for the proposed DisenGCD to adopt the GAT to learn the representations.

\section{Limitation Discussion}
There are still some limitations in the proposed  DisenGCD.
\begin{itemize}
    \item \textbf{High Computational Complexity and Poor Scalability.} The complexity of DisenGCD consists of three parts: the aggregation of $P$-hypernode meta multigraph on  $N+M+K$-node interaction graph (average $A_1$ neighbors),  L-layer GAT aggregation on the  $M+K$-node relation graph (average $A_2$ neighbors), and  L-layer GAT aggregation on the  $K$-node dependency graph (average $A_3$ neighbors). 
Suppose each node is  $d$ dimensional,
L-layer GAT on   $K$-node graph contains: 
computing  nodes' attention from  neighbors $O(L\times K\times A_3\times d)$ and   nodes'  linear transformation $O(L\times K \times d^2)$, totally equalling $O(L\times K\times d \times ( A_3+ d))$;
meta multigraph on $N+M+K$-node graph contains:
computing  nodes' attention $O(P\times (N+M+K)\times A_1\times d)$, 
computing path's attention $O(P\times (N+M+K)\times d)$, and nodes'  linear transformation
$O(P\times (N+M+K)\times d^2)$.
As a result, DisenGCD's complexity equals $O(P(N+M+K) d(A_1+1+d))+O(L(M+K)\times d(A_2+d))+O(L\times K\times d (A_3+d))$. 
As analyzed above, the time complexity of our model may be high and not applicable in some large data sets.

\item \textbf{Potentially Poor Task Transferability.} The proposed method primarily targets cognitive diagnosis tasks and is designed to handle such graphs as the \textit{student-exercise-concept} graph, which may not be readily applicable to other tasks. 
It is specifically tailored for modeling tasks similar to the student-exercise-concept triad diagram.
\end{itemize}

\end{appendices}

\newpage

\end{document}